\pdfoutput=1

\documentclass[11pt]{article}

\usepackage[preprint]{acl}

\usepackage{times}
\usepackage{latexsym}
\usepackage[T1]{fontenc}

\usepackage{inconsolata}

\usepackage[utf8]{inputenc} 
\usepackage[T1]{fontenc}    
\usepackage{hyperref}       
\usepackage{url}            
\usepackage{booktabs}       
\usepackage{amsfonts}       
\usepackage{nicefrac}       
\usepackage{microtype}      
\usepackage{xcolor}         
\usepackage{graphicx} 
\usepackage{colortbl}       
\usepackage{float}
\usepackage{multirow}
\usepackage{natbib}
\usepackage{algorithm}
\usepackage{algorithmic}
\usepackage{amsmath}
\usepackage{graphicx}
\usepackage{subcaption}
\usepackage{arydshln}
\usepackage{lipsum}
\usepackage{enumitem}
\setlist[enumerate]{nosep}

\title{K-Level Reasoning: Establishing Higher Order Beliefs \\in Large Language Models for Strategic Reasoning}

\author{
 \textbf{Yadong Zhang\textsuperscript{1,2,}\thanks{Work was done when interning at Microsoft Research Asia. † Correspondence to: shaoguang.mao@microsoft.com}},
 \textbf{Shaoguang Mao\textsuperscript{2,†}},
 \textbf{Tao Ge\textsuperscript{2}},
 \textbf{Xun Wang\textsuperscript{2}},
\\
 \textbf{Yan Xia\textsuperscript{2}},
 \textbf{Man Lan\textsuperscript{1}},
 \textbf{Furu Wei\textsuperscript{2}},
\\
\\
 \textsuperscript{1}East China Normal University,
 \textsuperscript{2}Microsoft Research Asia
\\
}

\begin{document}
\maketitle
\begin{abstract}
Strategic reasoning is a complex yet essential capability for intelligent agents. It requires Large Language Model (LLM) agents to adapt their strategies dynamically in multi-agent environments. Unlike static reasoning tasks, success in these contexts depends on anticipating other agents' beliefs and actions while continuously adjusting strategies to achieve individual goals. LLMs and LLM agents often struggle with strategic reasoning due to the absence of a reasoning framework that enables them to dynamically infer others' perspectives and adapt to changing environments.
Inspired by the Level-K framework \footnote{According to \href{https://en.wikipedia.org/wiki/Cognitive_hierarchy_theory\#The_Level-k_Framework}{the Level-k Framework}, k-level thinking involves considering what opponent/partner are likely to do, what they think you will do, and what they believe you think they will do, and so on.} from game theory and behavioral economics, which extends reasoning from simple reactions to structured strategic depth, we propose a novel framework: "\textbf{K-Level Reasoning with Large Language Models (K-R)}." This framework employs recursive mechanisms to enable LLMs to achieve varying levels of strategic depth, allowing agents to form higher order beliefs—beliefs about others' beliefs. We validate this framework through rigorous testing on four testbeds: two classical game theory problems and two social intelligence tasks. The results demonstrate the advantages of K-R in strategic reasoning. Our work presents the first recursive implementation of strategic depth in large language models (LLMs). It establishes a foundation for future research into theory of mind and strategic reasoning in LLMs.

\end{abstract}

\section{Introduction}
Strategic reasoning—decision-making in multi-participant environments—presents unique challenges for Large Language Models (LLMs) and LLM agents\cite{zhang2024llm}. In these settings, agents must respond to the actions of others while adapting to dynamic environments. They also need to align their decisions with their own goals during these interactions. Strategic reasoning is essential for intelligent agents and is widely applied in real-world tasks, such as investment, business strategy making\cite{zhao2023competeai}, negotiation\cite{hua2023war}, and policy-making\cite{li-etal-2024-econagent}.

\begin{figure}[t]
\centering
  \includegraphics[width=0.75\columnwidth]{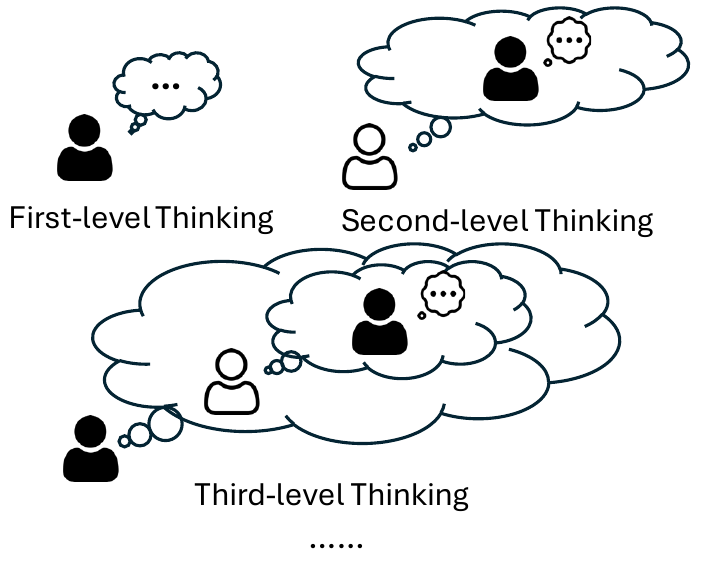}
  \caption{Level-K Framework: In first-level thinking, agents respond directly to the environment. In second-level thinking, agents consider the first-level thinking of others.  This process continues iteratively, with agents forming higher order beliefs based on assumptions about others' thoughts.}
  \label{fig:k-thinking}
  \vspace{-0.8cm}
\end{figure}

Effective strategic reasoning relies on understanding others' perspectives and anticipating their strategies. While there are some research efforts on LLMs' strategic reasoning, most methods rely on static prompting \cite{fu2023improving, xu2023exploring}. This typically involves instructing the model to account for others' beliefs and decisions during its own decision-making process in the prompt. However, these approaches fall short in enabling LLMs to form true higher order beliefs—beliefs about what others believe, and lack the flexibility needed for deeper strategic reasoning. 

\begin{figure*}[!ht]
  \centering
\includegraphics[width=\textwidth]{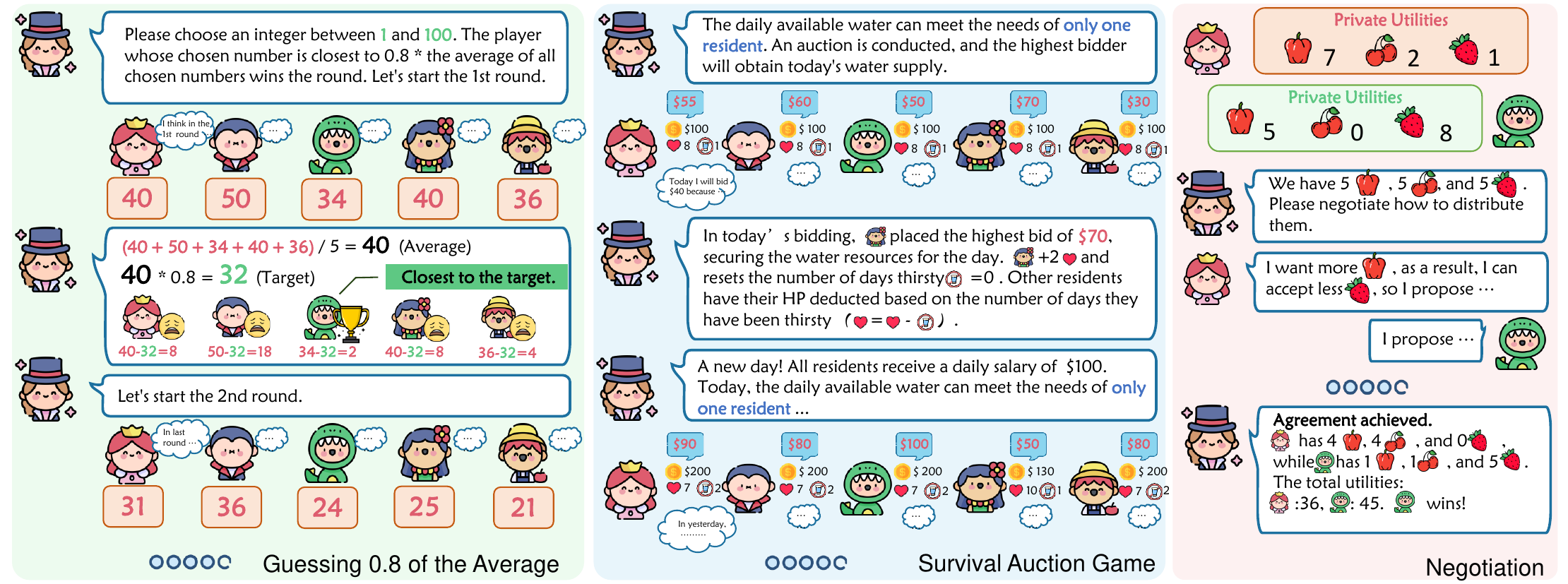} 
\caption{The illustration of three reasoning problems in dynamic, interactive environments in this paper. Left: Guessing 0.8 of the Average; Middle: Survival Auction Game; Right: Negotiation.}
\vspace{-0.2cm}
\label{fig:game_illustration}
\end{figure*}
K-level thinking (Figure \ref{fig:k-thinking}) \cite{nagel1995unraveling, cui2021k}, a classical concept in behavioral economics and game theory, categorizes reasoning into varying depths of strategic thought. It involves not only predicting others' actions but also considering their beliefs about one's actions, and even further layers of recursive thinking. 

Inspired by K-level thinking, we propose a novel strategic reasoning framework termed "K-Level Reasoning with LLMs (K-R)." K-R organizes reasoning into hierarchical levels and employs a recursive mechanism to integrate varying strategic depth into decision-making. Specifically, it involves: 1) recursively anticipating others' actions at varying levels of strategic depth with environmental context and historical public information, and 2) reasoning the optimal action based on these anticipations. To the best of our knowledge, this is the first approach to implementing varying levels of strategic depth in LLMs using a recursive mechanism and enables deeper reasoning in LLM agents through an algorithmic framework.

We validate this framework through rigorous testing on four testbeds: two classical game theory problems and two social intelligence tasks. The game theory problems includes Guessing 0.8 of the Average (Figure \ref{fig:game_illustration} left) and Survival Auction Game \cite{mao2023alympics} (Figure \ref{fig:game_illustration} middle). The social intelligence tasks includes Negotiation \cite{cao2018emergent} (Figure \ref{fig:game_illustration} right) and SOTOPIA benchmark\cite{zhou2023sotopia}. 
These settings serve as microcosms of the complex decision-making processes involved in strategic reasoning. Through extensive experiments, we demonstrate that our framework significantly outperforms existing reasoning methods and flexibly achieves varying levels of strategic depth. In addition to empirical evidence, we provide a theoretical analysis highlighting the benefits of K-R. We show that, leveraging the in-context learning capabilities of LLMs,
K-R can effectively model opponents' behavior using accumulated public and available opponent information.

Furthermore, we align the strategic depth of LLMs with human participants \cite{nagel1995unraveling, bosch2002one}. Using human as anchors, we observe that the K-R significantly enhances the strategic depth of LLMs from 0.25 to 1.89. Notably, when K=3, the strategic depth (1.89) of the LLM closely approaches that of financial newspaper readers (1.91). This strongly indicates that K-R establishes higher order beliefs in LLMs for strategic reasoning.

The contributions of this work are as follows:
\begin{itemize} [noitemsep]
\item We introduce K-R, a novel framework that extends k-level thinking to LLMs, enabling flexible strategic reasoning at varying depths through a recursive mechanism.

\item We conduct extensive evaluations, including game theory and social intelligence problems, demonstrating that K-R significantly outperforms existing methods in terms of flexibility and effectiveness, across both closed-source and open-source models.

\item We provide an in-depth analysis of K-R, confirming its ability to build higher order beliefs and enhance strategic reasoning. This lays a foundation for future research in theory of mind and strategic reasoning in LLMs.
\end{itemize}

\section{K-Level Reasoning with Large Language Models}
\label{sec:method}
\subsection{Methodology}
\label{sec:method_imp}
Strategic reasoning requires considering both the decision context and the possible actions of other participants.
We employ a multi-round normal form multi-participant game to introduce the proposed method. In this setting, an agent's decision-making process is formalized as follows: each agent \(i\) selects an action \(a_i^t\) from a set \(A_i^t\) at timestep \(t\). The payoff for agent \(i\), resulting from the collective action profile \(\mathbf{A}^t = (a_1^t, a_2^t, ..., a_N^t)\) and environment $E^t$, is denoted as \(U_i(E^t, \mathbf{A}^t)\).

At \(k=1\), agents decide based on environment $E^t$ without strategic anticipation:
\begin{equation}
    a_i^{t,1} = \arg \max_{a_i \in A_i^t}  \mathbb{E}[U_i(E^t, a_i)]
\end{equation}
At higher level thinking (\(k \ge 2\)), agent \(i\) simulates other agents operating at level \(k-1\) and adjusts their strategy accordingly\footnote{\space To simplify the formulation, we assume that all opponents are in the same thinking level. In practice, varying thinking level can be implemented.}: 
\begin{equation}
    a_i^{t,k} = \arg \max_{a_i \in A_i^t} \mathbb{E}[U_i(E^t,a_i, \hat{a}_{-i}^{t,k-1})]
\end{equation}
where \( \hat{a}_{-i}^{t,k-1} \) are the predicted actions of other agents based on their \(k-1\) level reasoning.

We propose a novel \textbf{strategic reasoning} framework with \textbf{recursive mechanisms}, termed ``K-Level Reasoning with Large Language Models (K-R),'' involving 1) recursively anticipating the actions \(\hat{a}^{t,k}_{-i}\) of others at different thinking levels using environment contexts and historical public information, followed by 2) reasoning the optimal action \(a_i^{t,k}\) based on anticipation of others' actions.

The K-Level Reasoning process is formulated as follows:

1) \textbf{Anticipation:}
\begin{equation}
\begin{split}
\hat{a}^{t, m}_j = 
\begin{cases} 
\text{LLM}(E^t, H_j^t) & \text{if } m = 1  \\ 
\text{LLM}(E^t, H_j^t, \hat{a}^{t,m-1}_{-j}) & \text{if } m > 1  \label{eq:recursive} 
\end{cases}
\end{split}
\end{equation}
where $\mathcal{H}_j^t$ = \(\{(E^1, a_j^1), (E^2, a_j^2), ..., (E^{t-1}, a_j^{t-1})\}\) represents public historical data of agent $j$, and $m$ denotes the specified thinking level.

2) \textbf{Reasoning:}

   \begin{equation}
      a_i^{t,k} = \text{LLM}(E^t, H_i^t, \hat{a}^{t,k-1}_{-i})
   \end{equation}

Algorithm \ref{alg:klevel-reasoning} outlines the implementation of K-R. This recursive method enables flexible and progressively deeper strategic reasoning (\(1, 2, ..., k, k+1, ...\)), thereby enhancing higher order belief in LLM agents.

\begin{algorithm}[t]
    \caption{K-Level Reasoning with LLMs}
    \label{alg:klevel-reasoning}
    \begin{algorithmic}[1]
    \REQUIRE $E^t$: Current decision context at time $t$; $H^t_i$: Historical information up to time $t$ for agent $i$; 
         \\ \ \ \ \ \ \ \     
    $K$: Depth of strategic reasoning; 
    \ENSURE $a_i^{t,K}$: Action for agent $i$ at time $t$ after K-level reasoning.
        
        \STATE \textbf{Function} \textit{\textbf{K\_REASONING}}$(i, k)$:
        \IF{$k == 1$}
            \RETURN $\text{LLM}(E^t, H^t_i)$
        \ELSE
            \FOR{each agent $j \neq i$}
            \STATE $\hat{a}^{t, k-1}_j = \textit{\textbf{K\_REASONING}}(j, k-1)$
            \ENDFOR
            \RETURN $\text{LLM}(E^t, H^t_i, \{\hat{a}^{t, k-1}_j \mid j \neq i\})$
        \ENDIF
    \STATE $a_i^{t,K} =$ \textit{\textbf{K\_REASONING}}$(i, K)$
    
    \RETURN $a_i^{t,K}$
    \end{algorithmic}
\end{algorithm}

\subsection{Theoretical Analysis}
\label{sec:theory}
This section discusses the benefits from K-R from a theoretical perspective. We utilize the in-context learning capabilities of LLMs to effectively model opponents' behavior. Suppose agent \( j \)'s decision-making process follows a hidden strategy \(\theta_j^*\). Thus, agent \( j \)'s decision-making can be expressed as:
\begin{equation}
P(a_j^t \mid E^t, \theta_j^*)
\end{equation}

The in-context learning of LLMs can be formally defined as implicit Bayesian inference\cite{xie2021explanation}; therefore, given the environment \( E^t \), the next action prediction conditioned on \( H^t_j \) is:
\begin{equation}
P(a_j^t \mid E^t, H^t_j) = \int P(a_j^t \mid E^t, \theta_j) P(\theta_j \mid H^t_j) d\theta_j
\label{eqq:P(a|eh)}
\end{equation}
As \( t \to \infty \), by the law of large numbers and properties of Bayesian updating, the posterior distribution concentrates around the true parameter \(\theta_j^*\):
\begin{equation}
P(\theta_j \mid H^t_j) \to \delta(\theta_j - \theta_j^*)
\end{equation}
where \(\delta\) is the Dirac delta function. 
Therefore,
\begin{equation}
\int P(a_j^t \mid E^t, \theta_j) P(\theta_j \mid H^t_j) d\theta_j \to P(a_j^t \mid E^t, \theta_j^*)
\end{equation}

This implies that as the number of interactions increases, K-R can more accurately predict opponents' behavior. 

It is also worth noting that interaction data cannot be infinite, and in-context learning is related to the performance of large language models (LLMs). Therefore, we empircally validate these hypotheses and reasoning in Section \ref{sec:predict}.


\section{Experiments: Game Theory}
\label{sec:exp_got}
To fairly compare the strategic reasoning capabilities of LLMs, we first adopt two widely used game theory settings. These controlled, well-defined game theory problems provide a robust assessment of LLMs' performance, with detailed setups outlined in Appendix \ref{app:game_setting}.
\subsection{Task Definition and Metrics}
\subsubsection{Guessing 0.8 of the Average (G0.8A)}
\textbf{G0.8A} (Figure \ref{fig:game_illustration} Left) is a classic game theory problem introduced by Alain Ledoux \cite{ledoux1981concours}. It involves 10-round games where each player selects a number between 1 and 100. The objective is to choose a number closest to 80\% of the group's average choice. The key idea is to guess how others will estimate the average and decide the number to submit. This concept is also illustrated in the Keynesian Beauty Contest \cite{keynes1936general}. This game mirrors the challenge of anticipating collective behavior in financial markets, where investors must predict not only the value of an asset but also how others will value it in the future.

The performance of the agent is evaluated using the \textbf{Win Rate}. Specifically, the Win Rate is calculated based on the wins achieved by the agent in individual round, rather than an entire game episode.

\subsubsection{Survival Auction Game (SAG)}
\textbf{SAG} (Figure \ref{fig:game_illustration} Middle) is derived from the Water Allocation Challenge proposed in \cite{mao2023alympics}. Each resident's goal is to survive a 10-day drought period by bidding for water resources and maintaining health points above zero. If a player successfully bids for water, they gain health points; otherwise, they lose health points. This integration of the auction system with the health points mechanism creates a dynamic environment where players must balance health and finances.

We use \textbf{Average Survival Round} measures the mean round in which a player remains active in the game.



\subsection{Base Techniques}
\label{sec:baselines}
We adapt a variety of approaches, originally from traditional reasoning and agent benchmarks. These base techniques include:

\textbf{Standard Prompting (Direct)}: This is the conventional prompting method in which the LLM generates the final answer (Action) in response to the given game setting prompt.

\textbf{Chain-of-Thought (CoT)}   \cite{wei2022chain}: We employ the zero-shot Chain-of-Thought reasoning method   \cite{kojima2022large}.

\textbf{Persona Prompting (Persona)}   \cite{deshpande2023toxicity}: This technique modifies the standard prompting process by incorporating a ``Game Expert'' persona to enhance the reasoning capabilities of the LLM.

\textbf{Reflexion} (\textbf{Reflect})   \cite{shinn2023reflexion}: This method refers to language agents with verbal reinforcement learning and has been adapted for dynamic tasks. Detailed modifications are explained in \ref{app:reflect}.

\textbf{Self-Refine} (\textbf{Refine})   \cite{madaan2023self}: This is a multi-round iterative reasoning approach where an additional LLM offers comments and adjustments prior to reaching a final decision. The distinctions between Self-Refine and Reflect are elaborated upon in the Appendix \ref{app:time}.

\textbf{Prediction Chain of Thought (PCoT)}: This strong baseline diverges from CoT by requiring the LLM to explicitly predict opponents' actions before making decisions. Unlike K-Level Reasoning, which involves a recursive approach, PCoT focuses on direct prediction based on context.

For implementation details and specific examples, please refer to Appendix \ref{app:prompt}.

\subsection{Experimental Settings}
\label{sec:exp-setting}
We established a controllable environment and distinguished between two roles: the player (primary focus) and the opponents. The player is equipped with a specific method, while all opponents use another reasoning approach. This well-defined setting allows for a clearer comparison of reasoning capabilities between methods.

In \textbf{G0.8A} and \textbf{SAG}, there is \textbf{one player} and \textbf{four opponents} for each game. Experiments for each setting are repeated 10 times and have passed the significance test (Appendix \ref{app:significance}), and each experiment consists of a 10-round game.

All methods in main experiments were implemented using GPT-4 \cite{achiam2023gpt} (gpt4-32k), with the temperature set at 0.7 and the top-p set at 0.9. We also conducted experiments with open-source LLMs, Details of which are provided in Appendix \ref{app:kr-open-llm}. Unless specified otherwise, the level of thinking in K-Level Reasoning is set to K=2.

\subsection{Results}


To distinguish between ``Player'' and ``Opponent'' in the table, the headers for \textbf{Player} (bold) and \textit{Opponents} (italics) are formatted accordingly. 

\begin{table}[ht]
    \centering
    \begin{minipage}[t]{\linewidth}
        \centering
        \caption{Win Rate of the player against different opponents in G0.8A game.} 
        
        \begin{minipage}[c]{0.99\textwidth}
        \centering
        \tabcolsep=0.1cm
        \centering
        \resizebox{\textwidth}{!}{
        \begin{tabular}{@{}l|cccccccc@{}}
        \toprule
        & \textbf{Direct} 	 &\textbf{CoT}    	 &\textbf{Persona}	 &\textbf{Reflect}	 &\textbf{Refine} 	 &\textbf{PCoT}   	 &\textbf{K-R} \\ \midrule
        \textit{Direct}      	& \cellcolor[rgb]{0.97,0.98,0.97} 0.43	&\cellcolor[rgb]{0.77,0.85,0.70} 0.67	&\cellcolor[rgb]{0.81,0.88,0.76} 0.62	&\cellcolor[rgb]{0.89,0.93,0.86} 0.53	&\cellcolor[rgb]{0.97,0.98,0.97} 0.43	&\cellcolor[rgb]{0.82,0.88,0.77} 0.61	&\cellcolor[rgb]{0.65,0.77,0.53} \textbf{0.82} \\
        \textit{CoT}         	& \cellcolor[rgb]{0.94,0.48,0.46} 0.07	&\cellcolor[rgb]{0.98,0.87,0.87} 0.32	&\cellcolor[rgb]{0.99,0.92,0.92} 0.35	&\cellcolor[rgb]{0.95,0.59,0.57} 0.14	&\cellcolor[rgb]{0.96,0.72,0.70} 0.22	&\cellcolor[rgb]{0.96,0.97,0.94} 0.45	&\cellcolor[rgb]{0.81,0.87,0.74} \textbf{0.63} \\
        \textit{Persona}     	& \cellcolor[rgb]{0.93,0.45,0.42} 0.05	&\cellcolor[rgb]{0.99,0.95,0.95} 0.37	&\cellcolor[rgb]{0.98,0.83,0.82} 0.29	&\cellcolor[rgb]{0.93,0.45,0.42} 0.05	&\cellcolor[rgb]{0.99,0.95,0.95} 0.37	&\cellcolor[rgb]{0.94,0.54,0.52} 0.11	&\cellcolor[rgb]{0.95,0.97,0.93} \textbf{0.46} \\
        \textit{Reflect}     	& \cellcolor[rgb]{0.98,0.99,0.98} 0.42	&\cellcolor[rgb]{0.76,0.85,0.69} 0.68	&\cellcolor[rgb]{0.81,0.87,0.74} 0.63	&\cellcolor[rgb]{1.00,0.98,0.98} 0.39	&\cellcolor[rgb]{0.80,0.87,0.73} 0.64	&\cellcolor[rgb]{0.71,0.81,0.62} 0.74	&\cellcolor[rgb]{0.68,0.79,0.58} \textbf{0.78} \\
        \textit{Refine}      	& \cellcolor[rgb]{0.94,0.53,0.51} 0.10	&\cellcolor[rgb]{0.99,0.91,0.90} 0.34	&\cellcolor[rgb]{0.98,0.87,0.87} 0.32	&\cellcolor[rgb]{0.98,0.86,0.85} 0.31	&\cellcolor[rgb]{0.97,0.73,0.72} 0.23	&\cellcolor[rgb]{0.96,0.72,0.70} 0.22	&\cellcolor[rgb]{0.95,0.97,0.93} \textbf{0.46} \\
        \textit{PCoT}        	& \cellcolor[rgb]{0.93,0.42,0.39} 0.03	&\cellcolor[rgb]{0.97,0.98,0.96} 0.44	&\cellcolor[rgb]{0.90,0.93,0.87} 0.52	&\cellcolor[rgb]{0.96,0.70,0.69} 0.21	&\cellcolor[rgb]{0.91,0.94,0.88} 0.51	&\cellcolor[rgb]{0.88,0.92,0.84} 0.54	&\cellcolor[rgb]{0.62,0.75,0.50} \textbf{0.85} \\
        \textit{K-R}         	& \cellcolor[rgb]{0.93,0.43,0.41} 0.04	&\cellcolor[rgb]{0.95,0.61,0.59} 0.15	&\cellcolor[rgb]{0.95,0.59,0.57} 0.14	&\cellcolor[rgb]{0.93,0.43,0.41} 0.04	&\cellcolor[rgb]{0.95,0.64,0.62} 0.17	&\cellcolor[rgb]{0.95,0.59,0.57} 0.14	&\cellcolor[rgb]{0.90,0.93,0.87} \textbf{0.52} \\
        \midrule
        Average		& \cellcolor[rgb]{0.95,0.63,0.61} \parbox[c]{1cm}{\ \  0.16\\± 0.18} 
        & \cellcolor[rgb]{0.98,0.99,0.97} \parbox[c]{1cm}{\ \  0.32\\± 0.19} 	
        & \cellcolor[rgb]{0.99,0.99,0.99} \parbox[c]{1cm}{\ \  0.41\\± 0.18} 	
        & \cellcolor[rgb]{0.97,0.75,0.73} \parbox[c]{1cm}{\ \  0.24\\± 0.18} 	
        & \cellcolor[rgb]{0.99,0.95,0.95} \parbox[c]{1cm}{\ \  0.37\\± 0.17}
        & \cellcolor[rgb]{1.00,1.00,1.00} \parbox[c]{1cm}{\ \  0.40\\± 0.25}
        & \cellcolor[rgb]{0.79,0.87,0.73} \parbox[c]{1cm}{\ \  \textbf{0.65}\\± 0.17}  \\
        \bottomrule
        \end{tabular}}
        \label{tab:winrate}
        \end{minipage}
        \hfill
        \begin{minipage}[c]{0.0\textwidth}
        \centering
        \end{minipage}
    \end{minipage}
    \hfill
\end{table}
\begin{table}[ht]
    \centering
    \begin{minipage}[t]{\linewidth}
        \centering
        \caption{Average Survival Round of the player against different opponents in Survival Auction Game.}
        \label{tab:winrate-water}
        \begin{minipage}[c]{0.99\textwidth}
            \tabcolsep=0.1cm
            \centering
        \resizebox{\textwidth}{!}{
          \begin{tabular}{@{}l|cccccccc@{}}
        \toprule
         & \textbf{Direct} 	 &\textbf{CoT}    	 &\textbf{Persona}	 &\textbf{Reflect}	 &\textbf{Refine} 	 &\textbf{PCoT}   	 &\textbf{K-R} \\ \midrule
        \textit{Direct}      	& \cellcolor[rgb]{0.98,0.87,0.86} 5.90	&\cellcolor[rgb]{0.92,0.95,0.89} 7.00	&\cellcolor[rgb]{0.85,0.90,0.80} 7.50	&\cellcolor[rgb]{0.94,0.55,0.53} 4.70	&\cellcolor[rgb]{0.68,0.79,0.57} 8.70	&\cellcolor[rgb]{0.97,0.98,0.96} 6.60	&\cellcolor[rgb]{0.58,0.73,0.44} \textbf{9.40} \\
        \textit{CoT}         	& \cellcolor[rgb]{0.98,0.82,0.81} 5.70	&\cellcolor[rgb]{0.99,0.99,0.98} 6.50	&\cellcolor[rgb]{0.96,0.71,0.70} 5.30	&\cellcolor[rgb]{0.92,0.37,0.34} 4.00	&\cellcolor[rgb]{0.76,0.84,0.69} 8.10	&\cellcolor[rgb]{0.96,0.71,0.70} 5.30	&\cellcolor[rgb]{0.49,0.67,0.33} \textbf{10.00} \\
        \textit{Persona}     	& \cellcolor[rgb]{0.98,0.82,0.81} 5.70	&\cellcolor[rgb]{0.82,0.88,0.76} 7.70	&\cellcolor[rgb]{0.86,0.91,0.81} 7.40	&\cellcolor[rgb]{0.96,0.68,0.67} 5.20	&\cellcolor[rgb]{1.00,0.97,0.97} 6.30	&\cellcolor[rgb]{0.89,0.93,0.85} 7.20	&\cellcolor[rgb]{0.59,0.73,0.46} \textbf{9.30} \\
        \textit{Reflect}     	& \cellcolor[rgb]{0.58,0.73,0.44} 9.40	&\cellcolor[rgb]{0.58,0.73,0.44} 9.40	&\cellcolor[rgb]{0.51,0.68,0.35} 9.90	&\cellcolor[rgb]{0.96,0.68,0.67} 5.20	&\cellcolor[rgb]{0.69,0.80,0.59} 8.60	&\cellcolor[rgb]{0.75,0.84,0.67} 8.20	&\cellcolor[rgb]{0.49,0.67,0.33} \textbf{10.00} \\
        \textit{Refine}      	& \cellcolor[rgb]{1.00,0.97,0.97} 6.30	&\cellcolor[rgb]{1.00,1.00,1.00} 6.40	&\cellcolor[rgb]{0.76,0.84,0.69} 8.10	&\cellcolor[rgb]{0.93,0.45,0.42} 4.30	&\cellcolor[rgb]{0.75,0.84,0.67} \textbf{8.20}	&\cellcolor[rgb]{0.96,0.71,0.70} 5.30	&\cellcolor[rgb]{0.79,0.86,0.72} 7.90 \\
        \textit{PCoT}        	& \cellcolor[rgb]{0.70,0.81,0.61} 8.50	&\cellcolor[rgb]{0.55,0.71,0.41} 9.60	&\cellcolor[rgb]{0.51,0.68,0.35} \textbf{9.90}	&\cellcolor[rgb]{1.00,0.97,0.97} 6.30	&\cellcolor[rgb]{0.70,0.81,0.61} 8.50	&\cellcolor[rgb]{0.99,0.95,0.95} 6.20	&\cellcolor[rgb]{0.54,0.70,0.39} 9.70 \\
        \textit{K-R}         	& \cellcolor[rgb]{0.92,0.39,0.37} 4.10	&\cellcolor[rgb]{0.97,0.76,0.75} 5.50	&\cellcolor[rgb]{0.95,0.63,0.62} 5.00	&\cellcolor[rgb]{0.92,0.38,0.35} 4.04	&\cellcolor[rgb]{0.98,0.82,0.81} 5.70	&\cellcolor[rgb]{0.93,0.47,0.45} 4.40	&\cellcolor[rgb]{0.94,0.96,0.93} \textbf{6.80} \\
        \midrule
        Average		
        & \cellcolor[rgb]{0.98,0.99,0.98}  	\parbox[c]{1cm}{\ \  6.51\\± 1.82}
        & \cellcolor[rgb]{0.85,0.90,0.81}  	\parbox[c]{1cm}{\ \  7.44\\± 1.55}
        & \cellcolor[rgb]{0.83,0.89,0.78}  	\parbox[c]{1cm}{\ \  7.59\\± 1.95}
        & \cellcolor[rgb]{0.95,0.58,0.57}  	\parbox[c]{1cm}{\ \  4.82\\± 0.82}
        & \cellcolor[rgb]{0.81,0.88,0.75}  	\parbox[c]{1cm}{\ \  7.73\\± 1.21}
        & \cellcolor[rgb]{0.99,0.94,0.94}  	\parbox[c]{1cm}{\ \  6.17\\± 1.29}
        & \cellcolor[rgb]{0.63,0.76,0.52}   \parbox[c]{1cm}{\ \  \textbf{9.01}\\± 1.21}  \\
        \bottomrule
        \end{tabular}}
        \end{minipage}
        \hfill
        \begin{minipage}[c]{0.08\textwidth}
        \centering
        \end{minipage}
    \end{minipage}
\end{table}

Table \ref{tab:winrate} presents Win Rate of players utilizing different methods against various opponents in the \textbf{G0.8A} game. Notably, the K-R method demonstrates a superior Win Rate of 0.65, significantly exceeding the win rates of the other strategies. Table 2 provides insights into the Average Survival Round of players across different auction game strategies in \textbf{SAG}, with the K-R method again standing out. The K-R method achieves an average survival round of 9.01, considerably higher than all other methods.

The experiment result underscores the effectiveness of the K-R method in enhancing player strategy, suggesting its strategic superiority in the context of this game. Its effectiveness lies in its ability to anticipate opponent moves, outperforming other prompting methods.

The performance of Reflect did not demonstrate the effectiveness of the reasoning method. We hypothesize that this is due to the fact that, in dynamic environments, Reflect on the experiences summarized from the previous round \cite{shinn2023reflexion} may not be applicable to the subsequent round of the game. Furthermore, in both games, Refine did not show an advantage over CoT and was significantly lower than K-R. This is because Refine involves adjustments based on one's \textit{own} strategy. However, these adjustments do not explicitly consider the hidden strategies of the opponent's behavior, rendering them inapplicable against opponents employing different strategies.

\section{Experiments: Social Intelligence}
\label{sec:exp_si}
We then evaluate K-R in two social intelligence benchmarks to assess its performance in more open-ended realistic scenarios. Compared to the abstract and theoretical settings of Game Theory, these scenarios involve richer contextual backgrounds and complicated goal pursuits, which better demonstrate the value of LLM-based agents in practical applications, such as in chatbots and strategic decision making.
\subsection{Task Definition and Metrics}

\subsubsection{Negotiation (NEG) }
\textbf{NEG} (Figure \ref{fig:game_illustration} Right)\cite{cao2018emergent, duan2024gtbench} is an open-ended and realistic task.
In this setting, two agents are presented with three types of items: peppers, cherries, and strawberries. Each agent has private utility values for these items and must negotiate to allocate the public item pool. 

The agent who secures more utility upon reaching an agreement wins the game, and we calculated the \textbf{Win Rate} to assess the performance of different agents.


\subsubsection{SOTOPIA Benchmark}
\textbf{SOTOPIA }\cite{zhou2023sotopia} is an open-ended environment to simulate complex social interactions between artificial agents and evaluate their social intelligence. It includes a variety of social scenarios, and each scenario includes a context background, and private social goals of each agent. Meanwhile, each agent has a character profiles which consists of name, gender, personality, occupation,  etc. 

For each episode, agents are scored at the end of the interaction along each of seven dimensions in \textbf{SOTOPIA-Eval}, including Goal Completion (\textsc{Goal}), Believability (\textsc{Bel}), Knowledge (\textsc{Kno}), Secret (\textsc{Sec}), Relationship (\textsc{Rel}), Social Rules (\textsc{Soc}), Financial and Material Benefits (\textsc{Fin}).

\subsection{Experimental Settings}
\label{sec:soc-exp-setting}
We employed the majority of the reasoning approaches introduced in Section \ref{sec:baselines} as baseline models for comparison.

In \textbf{NEG}, the experiments followed the settings from \cite{cao2018emergent, duan2024gtbench}. There is \textbf{one player} and \textbf{one opponent} for each game. We test the performance 
of the Baselines and K-Level Reasoning in 100 repeated independent games. To eliminate positional advantages, we swapped the positions of each player for each setting. To ensure the reliability, three trials were conducted, and the results are reported as averages with standard deviation. 

Meanwhile, We adhered to the \textbf{SOTOPIA-hard} \cite{zhou2023sotopia} setup comprising a total of 100 episodes, which is commonly found to be challenging for LLMs, and utilize a fixed GPT-4o based agent as partner. Additionally, to evaluate the agents' scores, we utilized GPT-4 as the assessment model, as it has been determined by SOTOPIA benchmark \cite{zhou2023sotopia} to serve as a reliable serves as a proxy for human judgments in evaluating model performance across most dimensions and for human performance on the \textsc{Goal} dimension.

\subsection{Results}


\begin{table}[ht]
    \centering
    \tabcolsep=0.1cm
    \caption{Win Rate of the player against opponent in Negotiation Setting.}
    \label{tab:negotiation_main}
    \resizebox{0.5\textwidth}{!}{
\begin{tabular}{@{}l|ccccccc@{}}
    \toprule
    & \textbf{Direct} & \textbf{CoT} & \textbf{Persona} & \textbf{Reflect} & \textbf{Refine} & \textbf{PCoT} & \textbf{K-R} \\
    \midrule
    \textit{Direct}      &\cellcolor[rgb]{1.00,1.00,1.00} \underline{50.00} &\cellcolor[rgb]{0.89,0.93,0.85} 61.34 &\cellcolor[rgb]{1.00,0.99,0.99} 49.58 &\cellcolor[rgb]{0.83,0.89,0.78} 66.67 &\cellcolor[rgb]{0.84,0.90,0.79} 65.83 &\cellcolor[rgb]{0.87,0.91,0.83} 63.03 &\cellcolor[rgb]{0.79,0.86,0.72} \textbf{70.83} \\
    \textit{CoT}         &\cellcolor[rgb]{0.98,0.86,0.85} 38.66 &\cellcolor[rgb]{1.00,1.00,1.00} \underline{50.00} &\cellcolor[rgb]{0.98,0.83,0.82} 36.67 &\cellcolor[rgb]{0.99,0.95,0.95} 45.83 &\cellcolor[rgb]{0.99,0.95,0.94} 45.76 &\cellcolor[rgb]{1.00,0.97,0.96} 47.27 &\cellcolor[rgb]{0.95,0.96,0.93} \textbf{55.36} \\
    \textit{Persona}     &\cellcolor[rgb]{1.00,1.00,0.99} 50.42 &\cellcolor[rgb]{0.87,0.91,0.82} 63.33 &\cellcolor[rgb]{1.00,1.00,1.00} \underline{50.00} &\cellcolor[rgb]{0.80,0.87,0.73} 70.00 &\cellcolor[rgb]{0.82,0.88,0.77} 67.50 &\cellcolor[rgb]{0.87,0.92,0.83} 62.50 &\cellcolor[rgb]{0.79,0.86,0.72} \textbf{70.83} \\
    \textit{Reflection}  &\cellcolor[rgb]{0.97,0.79,0.78} 33.33 &\cellcolor[rgb]{0.96,0.97,0.94} 54.17 &\cellcolor[rgb]{0.97,0.75,0.74} 30.00 &\cellcolor[rgb]{1.00,1.00,1.00} 50.00 &\cellcolor[rgb]{0.93,0.95,0.90} \textbf{57.14} &\cellcolor[rgb]{0.95,0.97,0.93} \underline{55.00} &\cellcolor[rgb]{0.95,0.97,0.93} \underline{55.00} \\
    \textit{Refine}      &\cellcolor[rgb]{0.98,0.80,0.79} 34.17 &\cellcolor[rgb]{0.96,0.97,0.94} 54.24 &\cellcolor[rgb]{0.97,0.78,0.77} 32.50 &\cellcolor[rgb]{0.99,0.91,0.91} 42.86 &\cellcolor[rgb]{1.00,1.00,1.00} 50.00 &\cellcolor[rgb]{0.94,0.96,0.92} \textbf{55.77} &\cellcolor[rgb]{0.95,0.97,0.94} \underline{54.55} \\
    \textit{PCoT}        &\cellcolor[rgb]{0.98,0.84,0.83} 36.97 &\cellcolor[rgb]{0.97,0.98,0.96} 52.73 &\cellcolor[rgb]{0.98,0.84,0.84} 37.50 &\cellcolor[rgb]{0.99,0.94,0.93} 45.00 &\cellcolor[rgb]{0.99,0.93,0.92} 44.23 &\cellcolor[rgb]{1.00,1.00,1.00} \underline{50.00} &\cellcolor[rgb]{0.93,0.95,0.91} \textbf{57.00} \\
    \textit{K-R}         &\cellcolor[rgb]{0.97,0.74,0.73} 29.17 &\cellcolor[rgb]{0.99,0.93,0.93} 44.64 &\cellcolor[rgb]{0.97,0.74,0.73} 29.17 &\cellcolor[rgb]{0.99,0.94,0.93} 45.00 &\cellcolor[rgb]{0.99,0.94,0.94} \underline{45.45} &\cellcolor[rgb]{0.99,0.91,0.91} 43.00 &\cellcolor[rgb]{1.00,1.00,1.00} \textbf{50.00} \\
    \midrule
    {Average}     &\cellcolor[rgb]{0.98,0.86,0.85} \parbox[c]{1cm}{38.96 \\±2.53} 
    &\cellcolor[rgb]{0.96,0.97,0.94} \parbox[c]{1cm}{\underline{54.35} \\±0.50} 
    &\cellcolor[rgb]{0.98,0.85,0.84} \parbox[c]{1cm}{37.92 \\±5.84} 
    &\cellcolor[rgb]{0.98,0.99,0.97} \parbox[c]{1cm}{52.19\\±1.73} 
    &\cellcolor[rgb]{0.96,0.98,0.95} \parbox[c]{1cm}{53.70 \\±4.41} 
    &\cellcolor[rgb]{0.96,0.97,0.95} \parbox[c]{1cm}{53.80 \\±4.34} 
    &\cellcolor[rgb]{0.91,0.94,0.88} \parbox[c]{1cm}{\textbf{59.08} \\± 2.20} \\
    \bottomrule
\end{tabular}}
\end{table}

\begin{table}[ht]
    \tabcolsep=0.03cm
    \caption{SOTOPIA-Eval of the player against opponent in SOTOPIA-hard.}
    \label{tab:sotopia_main}
    \resizebox{0.5\textwidth}{!}{
\begin{tabular}{@{}l|cccc|cccc}
    \toprule
                 
&\textbf{Direct} & \textbf{CoT} & \textbf{Refine} & \textbf{K-R} & \textbf{Direct} & \textbf{CoT} & \textbf{Refine} & \textbf{K-R}  \\
Metric & \multicolumn{4}{c}{[GPT-4o]} 
& \multicolumn{4}{c}{[LLaMA-3.1-70B]} \\
 \midrule 
\textsc{Bel} [0–10]&8.97 & 9.00 & 9.00 & 8.97  & 8.88 & 8.85 & 8.90 & \textbf{8.97} \\
\textsc{Rel} [-5–5]&2.38 & 2.40 & 2.27 & \textbf{2.67}  & 1.38 & 1.18 & 0.82 & \textbf{2.40} \\
\textsc{Kno} [0–10]&6.05 & 6.05 & 6.25 & 6.25  & 5.88 & 5.53 & 5.33 & \textbf{6.12} \\
\textsc{Sec} [-10-0]&0.00 & -0.05 & 0.00 & 0.00 & -0.28 & -0.25 & -0.18 & \textbf{0.00} \\
\textsc{Soc} [-10–0]&-0.05 & 0.00 & -0.05 & 0.00 & -0.70 & -0.72 & -0.64 & \textbf{0.00} \\
\textsc{Fin} [-5–5]&\textbf{0.90} & 0.78 & 0.80 & 0.72  & 0.38 & 0.35 & -0.08 & \textbf{0.75} \\
\textsc{Goal} [0–10] & 6.35 & \textbf{6.60} & 6.15 & 6.47 & 5.35 & 5.40 & 4.95 & \textbf{6.38} \\
\midrule
Overall & \parbox{1.1cm}{\ \ 3.51 \\± 0.09} 
& \parbox[c]{1.1cm}{\ \ 3.54 \\± 0.08} 
& \parbox[c]{1.1cm}{\ \ 3.49 \\± 0.08} 
& \parbox[c]{1.1cm}{\ \ \textbf{3.59} \\± 0.09} 
& \parbox[c]{1.1cm}{\ \ 2.98 \\±   0.23} 
& \parbox[c]{1.1cm}{\ \ 2.90 \\±   0.26} 
& \parbox[c]{1.1cm}{\ \ 2.73 \\±   0.25} 
& \parbox[c]{1.1cm}{\ \ \textbf{3.52} \\±   0.13}\\
    \bottomrule
\end{tabular}}
\end{table}
The results presented in Table \ref{tab:negotiation_main} and Table \ref{tab:sotopia_main} illustrate the effectiveness of the K-Level Reasoning in the context of NEG and SOTOPIA-hard settings, respectively.

In NEG, the K-R method demonstrates a notable win rate of 59.08\%, positioning it significantly above the average win rates achieved by other methods. This indicates that, in most cases, the proposals generated through K-Level Reasoning are more advantageous to itself, as well as suggesting a tendency to accept the opponent's proposals when the perceived benefits are substantial.

The results from SOTOPIA reveal several intriguing findings. Firstly, while K-R demonstrates some improvement compared to other methods, the results are not statistically significant. We hypothesize that this may be due to the inherent tendency of GPT-4 based models to assign higher scores to responses generated by GPT-4 based agents. Notably, we observed that employing agents based on LLaMA 3.1 70B with K-R can lead to significant performance enhancements. 
Meanwhile, the overall metrics indicate that K-R achieves performance levels comparable to those of the GPT-4 model, highlighting K-R's potential in the realm of social intelligence.
\section{Discussions}


\subsection{Does K-R Efficiently Establish a Higher Order Belief in LLMs?}
\begin{table}[ht]
    \tabcolsep=0.05cm
    \caption{Human performance in G2/3A.}
    \label{tab:g08a_human}
    \resizebox{\columnwidth}{!}{
        \begin{tabular}{@{}l|cccccc@{}}
            \toprule
            \textbf{Experiments} & \textbf{Lab} & \textbf{Classroom} & \textbf{Take-home} & \textbf{Theorists} & \parbox{1.7cm}{\textbf{Internet \\ Newsgroup}} & \textbf{Newspaper} \\
            \midrule
            Mean Choice          & 35.13        & 26.84              & 25.20              & 17.15              & 22.16                       & 23.08              \\
            \midrule
            Strategic Depth      & 0.87         & 1.53               & 1.68               & \textbf{2.63}               & 2.01                        & 1.91\\
            \bottomrule
        \end{tabular}
    }
    
\end{table}

\begin{table}[ht]
    \tabcolsep=0.1cm
    \caption{LLM performance in G0.8A in the first round.}
    \label{tab:g08a_llm}
    \resizebox{\columnwidth}{!}{
\begin{tabular}{@{}l|llllllll@{}}
            \toprule
\textbf{Method} & \textbf{Direct} & \textbf{CoT} & \textbf{Persona} & \textbf{Refine} & \textbf{Reflect} & \textbf{PCoT} & \textbf{KR{[}k=2{]}} & \textbf{KR{[}k=3{]}} \\
            \midrule
Mean Choice                  & 47.29           & 37.8         & 41.0             & 41.0            & 45.2             & 44.0          & 38.42                & 32.79                \\
\midrule
Strategic Depth              & 0.25            & 1.25         & 0.89             & 0.89            & 0.45             & 0.57          & 1.18                 & \textbf{1.89} \\               
            \bottomrule
\end{tabular}}
\end{table}

As a classic game theory issue, the G0.8A problem has garnered significant research interest across various disciplines. We reference the experimental results of the classic research among human participants \cite{nagel1995unraveling, bosch2002one} as anchor points and present the average decisions made by the K-Level Reasoning method (GPT-4) in the first round. Through this comparison, we can observe the relative relationship between human cognitive levels and LLMs under different reasoning methods. 
The specific calculation method on strategic depth is described in Appendix \ref{app:metric}. The performance of humans and LLMs is shown in the Table \ref{tab:g08a_human} and Table \ref{tab:g08a_llm}

From these observations, we can conclude that 
even when employing SOTA models, the strategic depth of GPT-4 under Direct Prompt (0.25) cannot compete with that of lower-strategic-capability undergraduate students in laboratory settings (0.87). Furthermore, the K-Level reasoning approach significantly enhances the reasoning depth of large language models, increasing it from 0.25 to 1.89, and the strategic depth of the large language model (1.89) approaches that of a group of financial newspaper readers (1.91) when K=3.

\subsection{K-Level Reasoning Leads to More Accurate Predictions About Opponents}
\label{sec:predict}

\begin{figure}[ht]
\centering
  \includegraphics[width=1\columnwidth]{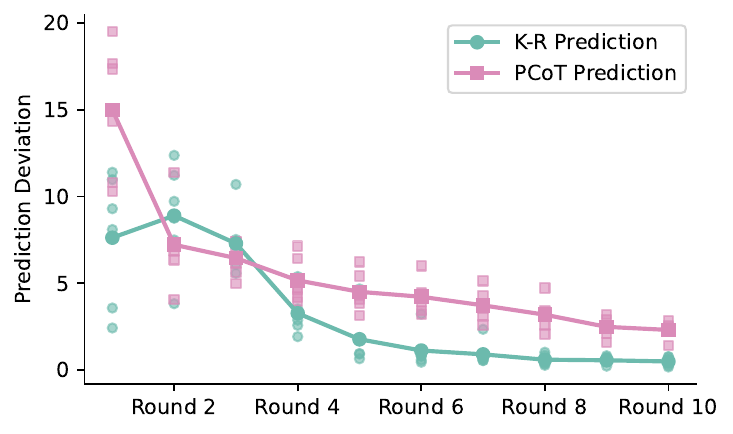}
\caption{The deviation in prediction during the G0.8A between PCoT and K-Level Reasoning.}
\label{fig:predict_error}
\end{figure}

Since K-R involves an intermediate step of modeling the opponent's behavior, we examine the progression of prediction accuracy. Figure \ref{fig:predict_error} illustrates the prediction deviation between K-R and PCoT in G0.8A. K-R exhibits higher prediction accuracy than PCoT from Round 1, starting with more precise and less random predictions. Moreover, the predictions converge quickly and become highly accurate in the second half of the game. This trend highlights the LLM's increasing proficiency in understanding higher order belief with more gameplay context. 
Essentially, K-R instantiates new sessions to compute the opponent's future actions. This approach leverages the in-context learning capabilities of LLMs more effectively than PCoT's prediction process (as theoretically discussed in Section \ref{sec:theory}). As a result, K-R achieves better prediction accuracy.

\subsection{Better Reasoning Methodology vs. Stronger Foundation Model}

There is a consensus that LLMs trained with more data and possessing larger parameter sizes demonstrate stronger reasoning capabilities. We explore whether K-Level Reasoning can significantly enhance the strategic reasoning abilities of relatively weaker LLMs. To investigate, we conducted experiments comparing the performance of K-R with GPT-3.5 (K-R[GPT-3.5]) against other reasoning methods based on GPT-4. All experiments were repeated 10 times.

\begin{table}[ht]
    \centering
        \tabcolsep=0.05cm
    \caption{A comparison of K-Level Reasoning with GPT-3.5 and other reasoning approaches with GPT-4. For the Guessing 0.8 of the Average, we report the win rate; for the Survival Auction Game, we report the average survival round.}
\label{tab:3.5vs4}
\resizebox{\linewidth}{!}{
    \begin{tabular}{@{}l|cccc|cccc@{}}
\toprule
& \multicolumn{4}{c|}{\textbf{ \textit{Guessing 0.8 of the Average}}}      & \multicolumn{4}{c}{\textbf{ \textit{Survival Auction Game}}} \\
\midrule
Opponent & \textbf{Direct}	& \textbf{K-R} & \textbf{Direct}	& \textbf{K-R} & \textbf{Direct}	& \textbf{K-R}	& \textbf{Direct} & \textbf{K-R} \\

[GPT-4]  	
& \multicolumn{1}{c}{\small [GPT-3.5]} 
& \multicolumn{1}{c}{\small [GPT-3.5]} 
& \multicolumn{1}{c}{\small [GPT-4]}
& \multicolumn{1}{c|}{\small [GPT-4]}
& \multicolumn{1}{c}{\small [GPT-3.5]} 
& \multicolumn{1}{c}{\small [GPT-3.5]}
& \multicolumn{1}{c}{\small [GPT-4]} 
& \multicolumn{1}{c}{\small [GPT-4]}\\

\midrule
\textit{Direct} 	& \cellcolor[rgb]{0.96,0.65,0.63} 0.18	& \cellcolor[rgb]{0.96,0.65,0.64} 0.18 	& \cellcolor[rgb]{0.97,0.98,0.97} 0.43 	& \cellcolor[rgb]{0.65,0.77,0.53} 0.82 & \cellcolor[rgb]{0.95,0.63,0.62} 5.00	& \cellcolor[rgb]{0.58,0.73,0.44} 9.40 	& \cellcolor[rgb]{0.98,0.87,0.86} 5.90 	& \cellcolor[rgb]{0.58,0.73,0.44} 9.40 \\
\textit{CoT}   & \cellcolor[rgb]{0.95,0.59,0.57} 0.14 		& \cellcolor[rgb]{0.99,0.95,0.95} 0.37 	& \cellcolor[rgb]{0.94,0.48,0.46} 0.07 	& \cellcolor[rgb]{0.81,0.87,0.74} 0.63 & \cellcolor[rgb]{0.96,0.71,0.70} 5.30	& \cellcolor[rgb]{0.76,0.84,0.69} 8.10 	& \cellcolor[rgb]{0.98,0.82,0.81} 5.70 	& \cellcolor[rgb]{0.49,0.67,0.33} 10.00 \\
\textit{Persona}	& \cellcolor[rgb]{0.94,0.53,0.51} 0.10	& \cellcolor[rgb]{0.97,0.73,0.72} 0.23 	& \cellcolor[rgb]{0.93,0.45,0.42} 0.05 	& \cellcolor[rgb]{0.95,0.97,0.93} 0.46 & \cellcolor[rgb]{0.95,0.63,0.62} 5.00	& \cellcolor[rgb]{0.85,0.90,0.80} 7.50 	& \cellcolor[rgb]{0.98,0.82,0.81} 5.70 	& \cellcolor[rgb]{0.59,0.73,0.46} 9.30 \\
\textit{Reflect}	& \cellcolor[rgb]{0.97,0.75,0.74} 0.24	& \cellcolor[rgb]{1.00,0.97,0.97} 0.38 	& \cellcolor[rgb]{0.98,0.99,0.98} 0.42 	& \cellcolor[rgb]{0.68,0.79,0.58} 0.78 & \cellcolor[rgb]{0.95,0.63,0.62} 5.00	& \cellcolor[rgb]{0.70,0.81,0.61} 8.50 	& \cellcolor[rgb]{0.58,0.73,0.44} 9.40 	& \cellcolor[rgb]{0.49,0.67,0.33} 10.00 \\

\textit{Refine}  & \cellcolor[rgb]{0.95,0.59,0.57} 0.14		& \cellcolor[rgb]{0.95,0.57,0.56} 0.13 	& \cellcolor[rgb]{0.94,0.53,0.51} 0.10 	& \cellcolor[rgb]{0.95,0.97,0.93} 0.46 & \cellcolor[rgb]{0.96,0.66,0.64} 5.10	& \cellcolor[rgb]{0.96,0.97,0.94} 6.70 	& \cellcolor[rgb]{1.00,0.97,0.97} 6.30 	& \cellcolor[rgb]{0.79,0.86,0.72} 7.90 \\
\textit{PCoT}   & \cellcolor[rgb]{0.96,0.67,0.65} 0.19		& \cellcolor[rgb]{0.95,0.97,0.93} 0.46 	& \cellcolor[rgb]{0.93,0.42,0.39} 0.03 	& \cellcolor[rgb]{0.62,0.75,0.50} 0.85 & \cellcolor[rgb]{0.92,0.39,0.37} 4.10 	& \cellcolor[rgb]{0.94,0.96,0.93} 6.80 	& \cellcolor[rgb]{0.70,0.81,0.61} 8.50 	& \cellcolor[rgb]{0.54,0.70,0.39} 9.70 \\
\midrule
Average	 & \cellcolor[rgb]{0.95,0.63,0.61} 0.16	& \cellcolor[rgb]{0.98,0.83,0.82} 0.29 	& \cellcolor[rgb]{0.96,0.66,0.64} 0.18 	& \cellcolor[rgb]{0.78,0.85,0.70} 0.67 & \cellcolor[rgb]{0.95,0.61,0.59} 4.92	& \cellcolor[rgb]{0.80,0.87,0.73} 7.83 	& \cellcolor[rgb]{0.93,0.95,0.90} 6.92 	& \cellcolor[rgb]{0.58,0.73,0.45} 9.38 \\
\bottomrule
\end{tabular}}
\end{table}

From the results in Table \ref{tab:3.5vs4}, we observe that K-R[GPT-3.5] outperforms the standard prompting method of GPT-4 (Direct[GPT4]) from average performance. Furthermore, when competing against opponents using reasoning methods on GPT-4, K-R[GPT-3.5] demonstrates remarkable capabilities. K-R, with its excellent 
restoration of the rival's perspective, enhances the LLM's ability in competitive environments. Additionally, we compared the performance of the open-source model LLAMA2-7B with GPT-3.5/4 in Appendix \ref{app:kr-open-llm}, finding that K-R significantly enhances reasoning in interactive contexts across different LLMs.

\subsection{The Deeper Thinking Level, the Better Strategic Performance?}

\newcolumntype{C}[1]{>{\centering\arraybackslash}p{#1}}

\begin{table}[ht]

\tabcolsep=0.3cm
\centering
\caption{Comparison between K-Level Reasoning[K=2] and K-Level Reasoning[K=3] in the two games.} 
\label{tab:k+1vsk}

\resizebox{\columnwidth}{!}{
\begin{tabular}{@{}lccc|ccc@{}}
\toprule
& \multicolumn{3}{c|}{\textbf{ \textit{Guessing 0.8 of the Average}}}  & \multicolumn{3}{c}{\textbf{ \textit{Survival Auction Game}}} \\
\midrule
\multicolumn{1}{l|}{Opponent}   & \textbf{Direct}	& \textbf{K-R[K=2]} & \textbf{K-R[K=3]}  & \textbf{Direct}	& \textbf{K-R[K=2]} & \textbf{K-R[K=3]}\\
\midrule
\multicolumn{1}{l|}{\textit{Direct}} 		& \cellcolor[rgb]{0.97,0.98,0.97} 0.43 &\cellcolor[rgb]{0.65,0.77,0.53} \textbf{0.82} & \cellcolor[rgb]{0.69,0.80,0.59} 0.77 (-0.05) & \cellcolor[rgb]{0.98,0.87,0.86} 5.90 & \cellcolor[rgb]{0.58,0.73,0.44} \textbf{9.40} & \cellcolor[rgb]{0.58,0.73,0.44} \textbf{9.40} (+0.00)\\
\textit{K-R[K=2]} 	& \cellcolor[rgb]{0.93,0.43,0.41} 0.04 &\cellcolor[rgb]{0.90,0.93,0.87} 0.52 & \cellcolor[rgb]{0.83,0.89,0.78} \textbf{0.60} (+0.08) & \cellcolor[rgb]{0.92,0.39,0.37} 4.10 & \cellcolor[rgb]{0.94,0.96,0.93} 6.80 & \cellcolor[rgb]{0.73,0.83,0.65} \textbf{8.30} (+1.50) \\

\bottomrule
\end{tabular}}
\end{table}

K-R models opponents' thinking processes recursively. We examine how thinking level affect reasoning outcomes by comparing K-R[K=2] and K-R[K=3] in two games. 
The results, detailed in Table \ref{tab:k+1vsk}, reveal the impact of increased thinking level.  Against the Direct method (first-level thinking), K-R[K=3] showed a decreased win rate in G0.8A but maintained performance in SAG, suggesting possible overthinking. However, K-R[K=3] improved significantly against K-R[K=2] in both games. It suggests that the key factor in K-R is the relative depth of thought compared to the opponent.  A one-level deeper approach offers a strategic advantage, but advancing two levels may lead to diminishing returns due to over-anticipation. In interactive environments, identifying opponents' thinking levels is difficult. Adapting to varying levels and using K-Level Reasoning for deeper analysis is a valuable direction for future research.

Additionally, a higher thinking level with the recursive prompting implementation increases computational cost. The computational cost of K-R is thoroughly discussed in Appendix \ref{app:computationalcost}.

\section{Related Work}
\subsection{Reasoning with LLMs}
Large Language Models (LLMs) excel in diverse complex reasoning tasks, such as mathematical  \cite{miao2021diverse, patel2021nlp}, common sense  \cite{talmor2022commonsenseqa, bhakthavatsalam2021think}, and symbolic reasoning  \cite{srivastava2022beyond,suzgun2022challenging}. 
A notable reasoning approach involves breaking down complex questions into a series of intermediate steps, a technique known as the Chain-of-Thought (CoT) method  \cite{wei2022chain, kojima2022large}. Subsequently, some works have emerged to extend CoT, with innovations like Tree of Thought (ToT)   \cite{yao2023tree}, Graph of Thought (GoT)  \cite{besta2023graph} and Skeleton-of-thought  \cite{ning2023skeleton}. Besides, approaches like Self-Refine  \cite{madaan2023self} and Reflexion  \cite{shinn2023reflexion} enhance CoT's consistency by having LLMs review and refine their responses. Moreover, recent research has revealed that integrating persona information into LLMs significantly improves their reasoning processes  \cite{deshpande2023toxicity}. A series of studies  \cite{fu2023improving,wang2023unleashing} have been conducted to incorporate more persona information, aiming to enhance the rationality and knowledge ability of the LLM reasoning process. 
These methods have been applied to various static tasks, but have not been adequately evaluated in dynamic problems (multi-agent environment) to validate their efficacy in reasoning capabilities.

\subsection{Strategic Reasoning within Multiple Agent System}
Dynamic problems arise when multiple participants are involved in multi-round interactions. One key factor is the simultaneous interactions of multiple participants with the environment.  Unlike single-agent systems, multiple agent system (MAS) encounters a broader range of issues and challenges, as noted by \cite{wong2021deep}, including computational complexity  \cite{ding2020challenges}, nonstationarity  \cite{papoudakis2019dealing}, partial observability  \cite{mahajan2019maven, foerster2016learning}, and challenges in credit assignment  \cite{sunehag2017value}. Particularly, in the context of inference using LLMs, the nonstationarity of the environment poses a distinct challenge. 

Recently, research on LLMs in strategic reasoning has been conducted across various MAS including social behavior\cite{zhou2023sotopia, hua2023war}, economic simulations\cite{zhao2023competeai, li2023tradinggpt}, game theory\cite{duan2024gtbench, xu2023magic}, and game playing\cite{ma2023large, xu2023exploring}. 
To enhance the performance of LLMs in strategic reasoning scenarios, researchers have utilized the concepts of Theory of Mind (ToM) \cite{gandhi2023strategic, guo2023suspicion} and Reinforcement Learning \cite{xu2023language, zhang2024agent} to optimize the reasoning processes of LLMs. These approaches involve prompting LLMs to recognize the intricacies of strategic tasks, like our proposed Prediction Chain-of-Thought baseline. However, our experimental results indicate that this approach fails to establish a clear cognitive hierarchy necessary for recursive and deeper strategic thinking. 

\section{Conclusion }
\label{sec:limit}
This paper represents a significant stride in understanding and enhancing the strategic reasoning capabilities of LLMs. 
We propose ``K-Level Reasoning with LLMs.'' This innovative approach leverages recursive mechanisms to achieve varying thinking level  within LLMs, enabling them to engage in deeper  strategic thinking.
Through extensive experiments, we validate the advantage offered by this method. It establishes a foundation for future
research into theory of mind and strategic reasoning in LLMs.


\section{Limitations}
We validate the effectiveness of the K-Level Reasoning framework from two perspectives: game theory and social intelligence. While our experimental results provide substantial evidence supporting the framework's validity, further research is necessary to explore the performance of large language models (LLMs) in few-shot agent modeling \cite{he2016opponent} across various environments, strategic factors, and action sets.

Additionally, K-R predicts opponents' most likely behavior by initiating a new LLM inference session. The recursive mechanism employed to achieve varying levels of strategic depth inevitably increases computational cost. Appendix \ref{app:computationalcost} provides a detailed discussion on how K-R relates to this rise in computational cost and compares it across different reasoning methods. Despite the increased demands, K-R outperforms other methods with comparable computational costs.
\bibliography{custom}

\appendix
\section{Impact Statements}
\label{app:impact}
Our research introduces the K-Level Reasoning framework, designed to formulate strategies in dynamic, interactive and competitive scenarios by anticipating the reactions of adversaries, potentially users. Theoretically, this approach offers a novel perspective for understanding and optimizing decision-making processes. However, we recognize when the goal setting diverges from user interests, the application of K-Level Reasoning could result in manipulative behaviors by adapting to the predicted user's reactions. This risk is notably pronounced in scenarios designed to influence user decisions or behaviors, such as in recommendation systems, advertising placements, and content distribution on social media platforms.

Although K-Level Reasoning provides a potent powerful tool for strategic planning, interacting and reasoning, ethical considerations must be meticulously managed in its practical application. This ensures that the development and utilization of technology do not detrimentally impact user and societal interests. To this end, we advocate for heightened transparency, ensuring users have a comprehensive understanding and control over how their data is utilized.

\section{Game Setting}
\label{app:game_setting}

\subsection{Guessing 0.8 of the Average}

\textbf{Initial Setup}: For each round, each player select a number between 1 and 100. The objective is to select a number that is closest to 80\% of the group's average choice. Formally, each player $i$ chooses a number $n_i$, aiming for $n_i\approx0.8 \times \overline{n}$, where $\overline{n}$ is the average of all chosen numbers.

\textbf{Scoring and Continuation}:
 A player scores a point if his/her chosen number is closest to 80\% of the average number chosen by the group. If all players select the same number, no points are awarded for this round. Mathematically, the score for player $i$ in in round $t$ is given by $s^t_i$, which is 1 if $|n_i - 0.8 \times \overline{n}|$ is the minimum among all players, and 0 otherwise.
\begin{figure*}[htbp]
  \centering
\includegraphics[width=0.83\textwidth]{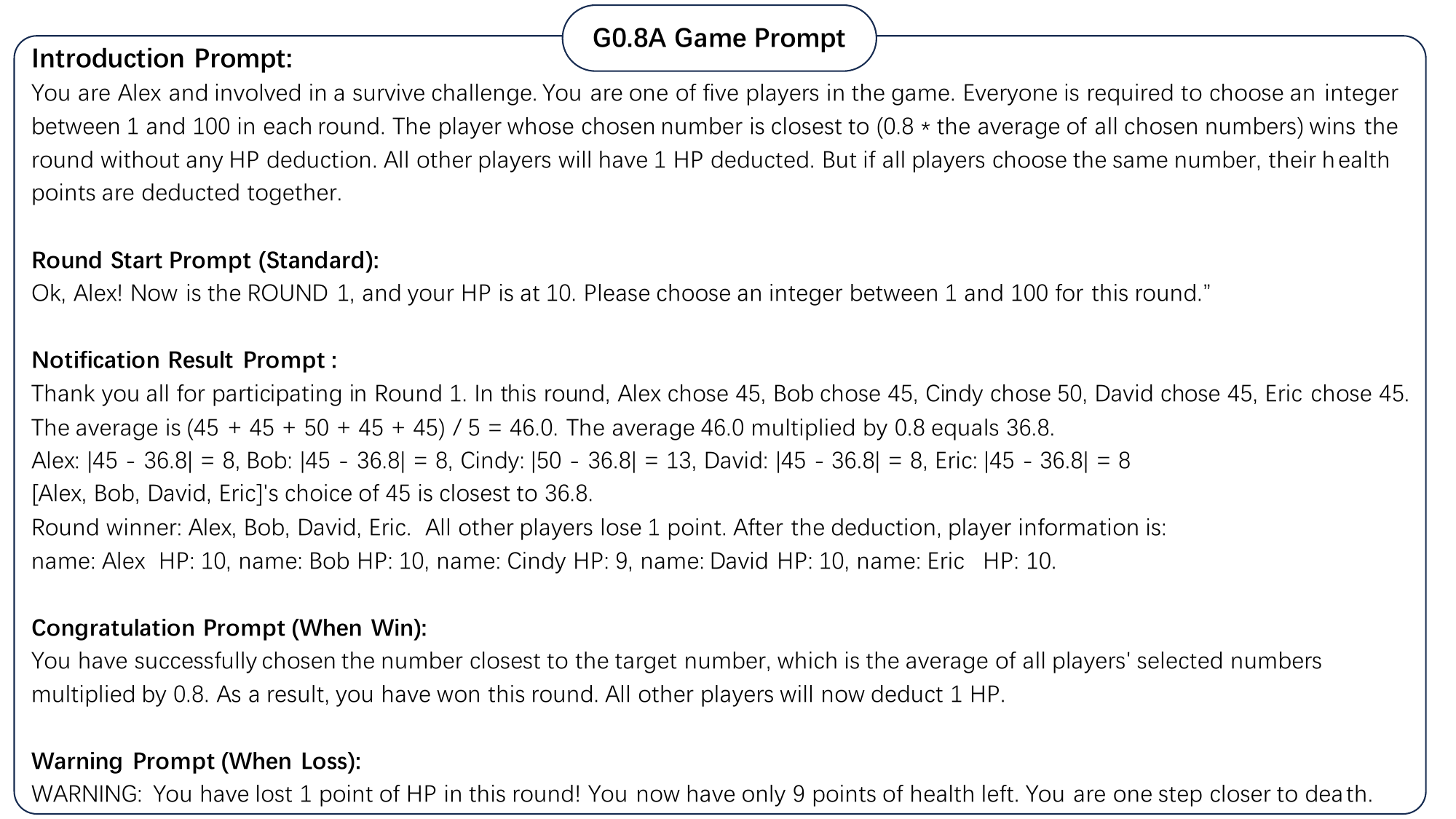}
  \caption{Prompts used in Guessing 0.8 of the Average game.}
\end{figure*}
\begin{figure*}[htbp]
  \centering
\includegraphics[width=0.83\textwidth]{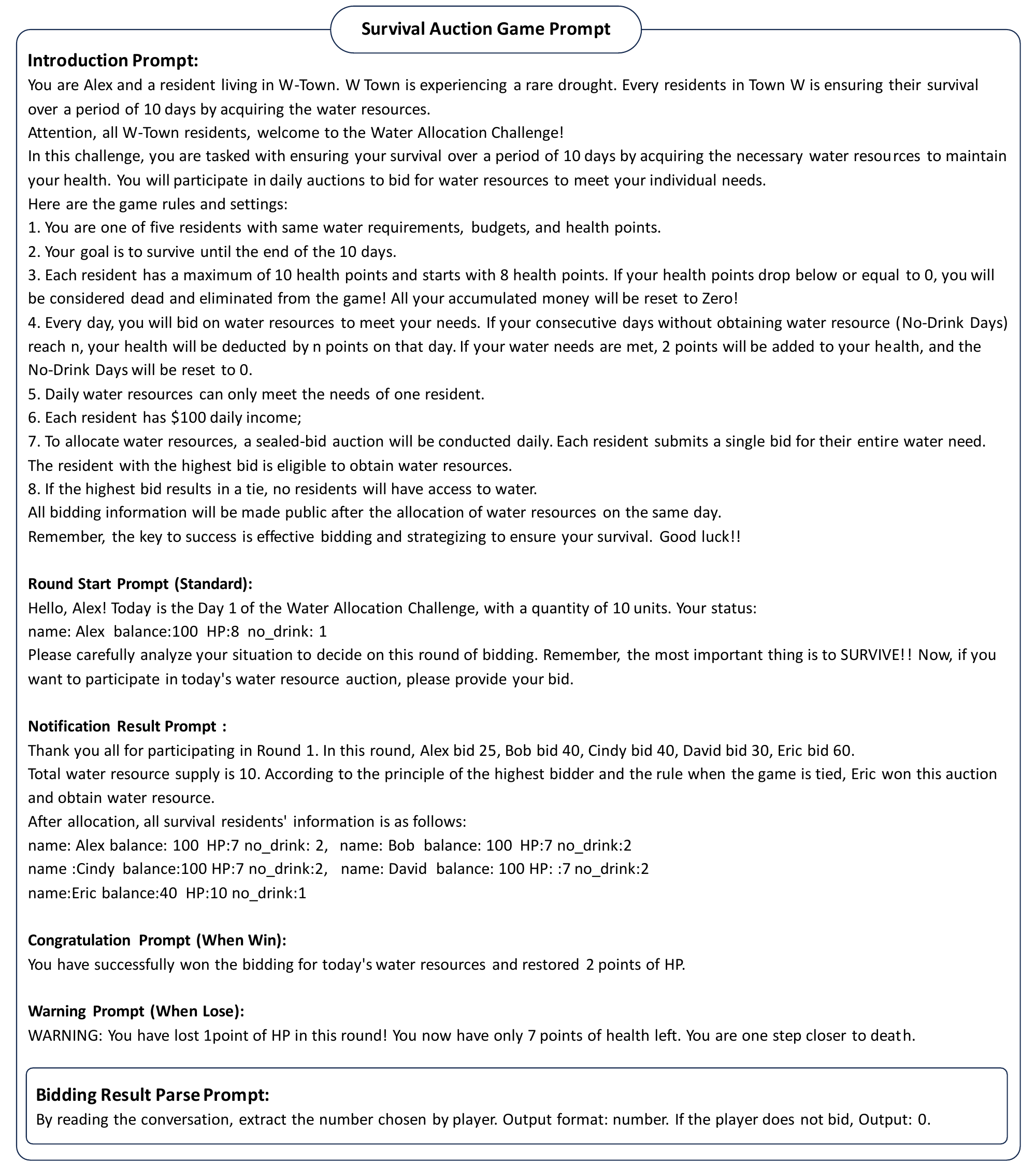}
  \caption{Prompts used in Survival Auction Game.}
\end{figure*}

\subsection{Survival Auction Game}

\textbf{Initial Setup}: Players start with 8 health points, out of a maximum of 10. Every day, each player possesses a fixed income of \$100. The daily water supply can only satisfy one resident's requirement. 

\textbf{Scoring and Continuation}:  Everyday, players engage in a daily auction to secure the necessary water resources, and the highest bidder wins. In case of a tie, the resources are not allocated to any player. If a player successfully bid the water resources, they will gain 2 health points; otherwise, they will lose health points equal to the number of consecutive days, denoted as $n$, during which they have not obtained water resources. Once a player's health points fall to 0 or below, they will be eliminated. The health point of player $i$ on day $t$, denoted as $h^t_i$, is crucial in determining their survival and bidding strategy.

\section{Detailed Metric Computational Formulas }
\label{app:metric}
\textbf{Win Rate} is 
calculated based on the number of wins 
over game going, providing a measure of the overall ability.
{
\small
\begin{equation}
\mathrm{WinRate} = \frac{\mathrm{Num\ of\ Wins} }{\mathrm{Total\ Round\ per\ Test} \times \mathrm{Num\ of\ Test} } 
\end{equation}
}

\textbf{Average Survival Round} calculates the average round in which the player remains in the game. It's an effective way to assess performance in elimination-based game, like SAG.
{
\small
\begin{equation}
\mathrm{AvgSurvivalRound} = \frac{\sum \mathrm{Survival\ Round\ in\ Each\ Test} }{\mathrm{Num\ of\ Test} } 
\end{equation}
}

\textbf{Prediction Accuracy} evaluates the accuracy of player's predictions regarding rivals' future moves. 
In the G0.8A, it involves calculating the absolute difference between the player's predicted average and the actual average in each round:
\begin{equation}
\mathrm{Pred\ Acc} = \frac{\sum|\mathrm{Avg}_{Pred}-\mathrm{Avg}_{Actual}|}{\mathrm{Num\ of\ Test}} 
\end{equation}
In the SAG, the focus shifts to measuring the absolute error between the player's prediction of the highest bid by opponents and the actual highest bid made by them.
\begin{equation}
\mathrm{Pred\ Acc} = \frac{\sum|\mathrm{Max\ Bid}_{Pred} - \mathrm{Max\ Bid}_{Actual}|}{\mathrm{Num\ of\ Test}}
\end{equation}

\textbf{Strategic Level}
\begin{equation}
    \mathrm{Strategic Depth}(\mathrm{choice})= \log_{\alpha}(\mathrm{choice}/50)
\end{equation}
Here, $\alpha$ represents the target value coefficient, and $50$ represents the average of a random choice between 0 and 100, which is used to represent level-0 players. In the settings of [1][2], the parameter  $\alpha$ is set to $\frac{2}{3}$. In our configuration,  $\alpha$  is set to $0.8$, which is the origin of the name G0.8A.

\section{Performance of Large Language Models Competing with Programmatic Strategies}
\label{app:LLMvsComputer}
In addition to using the \textbf{LLM-LLM Combat} comparison setting in Section 4, we have also designed a set of \textbf{LLM-Programmatic Strategy Combat }comparisons. In Player-Programmatic Strategy Combat setting, the ``player" will be equipped with a specific reasoning method, while opponents will play according to programmatic strategic patterns, which will not be adjusted with the game going. This mode is to check different methods' adaptation to different predefined but fixed patterns.
For the Player-Programmatic Player Combat Combat in G0.8A, the programmatic strategies include:

1) \textbf{0-Level (Fix)}: The characteristic of a 0-Level player is that their choice of strategy is uniform \cite{stahl1995players}. We limit the choice space of the 0-Level Computer player to only 40.

2) \textbf{0-Level (Var)}: Modified from the 0-Level (Fix) strategy, the selection is sampled from a Gaussian distribution with a mean of 40 and a variance of 5.

3) \textbf{MonoTrend (Fix)}: The numbers chosen by the computer players follow an arithmetic sequence with a decreasing common difference, and the common differences for all four computer players are the same.

4) \textbf{MonoTrend (Var)}: The numbers chosen by the computer players follow an arithmetic sequence with a decreasing common difference, and the common differences for the four computer players are randomly generated from 1 to 5.

5) \textbf{LastBids (Fix)}: The computer player chooses the target number from the previous round (selects 40 in the first round).

6) \textbf{LastBids (Var)}: Modified from the LastBids strategy, the selection is sampled from a Gaussian distribution with a mean equal to the target number of the previous round and a variance of 5.

Overall, the dynamic changes of these three settings are LastBids $>$ MonoTrend $>$ 0-Level. We use these three programs to test the reasoning capability to adapt to and counter different patterns. 

Table \ref{app:tab:prog_winrate} reveals a significant trend in the performance of players against other approaches. The effectiveness of reasoning approaches decreases in the following order: 0-Level, MonoTrend, and LastBids. This pattern highlights a reduction in the efficacy of the LLM in more dynamic environments. We found that only K-Level Reasoning shows an advantage in LastBids (Fix), indicating that compared to K-Level Reasoning, previous reasoning methods on static problems lack observation and judgment of the opponent. Conversely, this also demonstrates that K-Level Reasoning can implicitly infer the behavior pattern of the opponent based on their historical actions.

Intriguingly, reasoning methods significantly influence the performance in dynamic settings. Methods like CoT and Self-Refine, traditionally favored in static reasoning, also demonstrate substantial improvements over the Standard Prompt approach. This finding underscores the necessity for more elaborate reasoning processes in dynamic decision-making, akin to static problem-solving scenarios.

\begin{table}[ht]
\caption{Win Rate of the player against different programmatic strategies in Guessing 0.8 of the Average game.}
\begin{minipage}[c]{1\columnwidth}
\centering
\tabcolsep=0.1cm
\centering
\resizebox{\columnwidth}{!}{
\begin{tabular}{@{}lcccccccc@{}}
\toprule
Opponent & \textbf{Direct} 	 &\textbf{CoT}    	 &\textbf{Persona}	 &\textbf{Reflect}	 &\textbf{Refine} 	 &\textbf{PCoT}   	 &\textbf{K-R} \\ \midrule
\multicolumn{8}{@{}c}{\textbf{\textit{ Player VS Programmatic Strategies}}} \\[5pt]
\multicolumn{1}{l|}{\textit{0-Level (Fix)}}     	& \cellcolor[rgb]{0.79,0.86,0.72} 0.65	&\cellcolor[rgb]{0.60,0.74,0.48} 0.87	&\cellcolor[rgb]{0.60,0.74,0.48} 0.87	&\cellcolor[rgb]{0.65,0.77,0.54} 0.81	&\cellcolor[rgb]{0.50,0.68,0.34} 0.99	&\cellcolor[rgb]{0.66,0.78,0.56} 0.80	&\cellcolor[rgb]{0.52,0.69,0.37} 0.97 \\
\multicolumn{1}{l|}{\textit{0-Level (Var)}}	& \cellcolor[rgb]{0.97,0.98,0.96} 0.44	&\cellcolor[rgb]{0.77,0.85,0.70} 0.67	&\cellcolor[rgb]{0.76,0.84,0.68} 0.69	&\cellcolor[rgb]{0.82,0.88,0.77} 0.61	&\cellcolor[rgb]{0.88,0.92,0.84} 0.54	&\cellcolor[rgb]{0.70,0.80,0.60} 0.76	&\cellcolor[rgb]{0.69,0.80,0.59} 0.77 \\
\multicolumn{1}{l|}{\textit{MonoTrend (Fix)}}	& \cellcolor[rgb]{0.93,0.45,0.42} 0.05	&\cellcolor[rgb]{0.93,0.46,0.44} 0.06	&\cellcolor[rgb]{0.95,0.61,0.59} 0.15	&\cellcolor[rgb]{0.92,0.37,0.34} 0.00	&\cellcolor[rgb]{0.98,0.83,0.82} 0.29	&\cellcolor[rgb]{0.95,0.61,0.59} 0.15	&\cellcolor[rgb]{0.93,0.96,0.91} 0.48 \\
\multicolumn{1}{l|}{\textit{MonoTrend (Var)}}	& \cellcolor[rgb]{0.99,0.91,0.90} 0.34	&\cellcolor[rgb]{0.97,0.98,0.96} 0.44	&\cellcolor[rgb]{0.86,0.91,0.81} 0.57	&\cellcolor[rgb]{0.99,0.89,0.88} 0.33	&\cellcolor[rgb]{0.92,0.95,0.90} 0.49	&\cellcolor[rgb]{0.95,0.97,0.93} 0.46	&\cellcolor[rgb]{0.71,0.81,0.62} 0.74 \\
\multicolumn{1}{l|}{\textit{LastBids (Fix)}}    	& \cellcolor[rgb]{0.92,0.38,0.36} 0.01	&\cellcolor[rgb]{0.95,0.56,0.54} 0.12	&\cellcolor[rgb]{0.95,0.62,0.60} 0.16	&\cellcolor[rgb]{0.92,0.38,0.36} 0.01	&\cellcolor[rgb]{0.97,0.79,0.79} 0.27	&\cellcolor[rgb]{0.93,0.46,0.44} 0.06	&\cellcolor[rgb]{0.70,0.81,0.61} 0.75 \\
\multicolumn{1}{l|}{\textit{LastBids (Var)}}	& \cellcolor[rgb]{0.93,0.46,0.44} 0.06	&\cellcolor[rgb]{0.95,0.61,0.59} 0.15	&\cellcolor[rgb]{0.96,0.65,0.64} 0.18	&\cellcolor[rgb]{0.96,0.67,0.65} 0.19	&\cellcolor[rgb]{0.96,0.65,0.64} 0.18	&\cellcolor[rgb]{0.95,0.59,0.57} 0.14	&\cellcolor[rgb]{0.96,0.65,0.64} 0.18 \\
\bottomrule
\end{tabular}}
\label{app:tab:prog_winrate}
\end{minipage}
\end{table}

\section{Open Source LLM with K-Level Reasoning}
\label{app:kr-open-llm}

In addition to the experiments on GPT3.5 /GPT4, we conducted tests with an open-source smaller-scale model, LLaMA-7B-Chat, in the "G0.8A" game. To clearly compare the performance of different LLMs, we adopted the LLM-Programmatic Strategy Combat setting described in Appendix B. From the experimental results, we can see that: 1. Llama2-7B and GPT3.5, even including GPT4, perform poorly when competing against programmatic strategies using standard prompting, struggling even with some very simple strategies. It underscores the significance of our research; 2. For all tested models, applying k-level reasoning effectively enhanced the win rate, highlighting the stability of the improvement our proposed model brings to different base models. Interestingly, for stronger LLMs, applying K-R can achieve a higher relative improvement, we speculate that the enhancement is simultaneously derived from the opponent's action simulation and reasoning capabilities of the base model.

\begin{table}[ht]
\caption{Win Rate of the reasoning methods with open source LLMs against different programmatic strategies in Guessing 0.8 of the Average game.}
\begin{minipage}[c]{1\columnwidth}
\centering
\tabcolsep=0.1cm
\centering
\resizebox{\columnwidth}{!}{
\begin{tabular}{@{}lcccccc@{}}
\toprule
Methods & \textbf{Direct} 	 &\textbf{K-R[K=2]}    	 & \textbf{Direct} 	 &\textbf{K-R[K=2]} & \textbf{Direct} 	 &\textbf{K-R[K=2]} \\ \midrule
Base Model & [LLaMA-7B]	&[LLaMA-7B]	&[GPT3.5]	&[GPT3.5]	&[GPT4]	&[GPT4] \\ \midrule
\multicolumn{7}{@{}c}{\textbf{\textit{ Player VS Programmatic Strategies}}} \\[5pt]
\multicolumn{1}{l|}{\textit{0-Level (Fix)}}	& \cellcolor[rgb]{0.98,0.84,0.84} 0.30	&\multicolumn{1}{c|}{\cellcolor[rgb]{0.97,0.98,0.97} 0.43}	&\cellcolor[rgb]{0.98,0.84,0.84} 0.30	&\multicolumn{1}{c|}{\cellcolor[rgb]{0.92,0.95,0.90} 0.49}	&\cellcolor[rgb]{0.79,0.86,0.72} 0.65	&\cellcolor[rgb]{0.52,0.69,0.37} 0.97 \\
\multicolumn{1}{l|}{\textit{0-Level (Var)}}	& \cellcolor[rgb]{0.95,0.57,0.56} 0.13	&\multicolumn{1}{c|}{\cellcolor[rgb]{0.96,0.72,0.70} 0.22}	&\cellcolor[rgb]{0.94,0.54,0.52} 0.11	&\multicolumn{1}{c|}{\cellcolor[rgb]{0.99,0.94,0.93} 0.36}	&\cellcolor[rgb]{0.97,0.98,0.96} 0.44	&\cellcolor[rgb]{0.69,0.80,0.59} 0.77 \\
\multicolumn{1}{l|}{\textit{MonoTrend (Fix)}}	& \cellcolor[rgb]{0.95,0.56,0.54} 0.12	&\multicolumn{1}{c|}{\cellcolor[rgb]{0.95,0.56,0.54} 0.12}	&\cellcolor[rgb]{0.95,0.64,0.62} 0.17	&\multicolumn{1}{c|}{\cellcolor[rgb]{0.98,0.87,0.87} 0.32}	&\cellcolor[rgb]{0.93,0.45,0.42} 0.05	&\cellcolor[rgb]{0.93,0.96,0.91} 0.48 \\
\multicolumn{1}{l|}{\textit{MonoTrend (Var)}}	& \cellcolor[rgb]{0.95,0.64,0.62} 0.17	&\multicolumn{1}{c|}{\cellcolor[rgb]{0.95,0.57,0.56} 0.13}	&\cellcolor[rgb]{0.96,0.67,0.65} 0.19	&\multicolumn{1}{c|}{\cellcolor[rgb]{0.98,0.99,0.98} 0.42}	&\cellcolor[rgb]{0.99,0.91,0.90} 0.34	&\cellcolor[rgb]{0.71,0.81,0.62} 0.74 \\
\multicolumn{1}{l|}{\textit{LastBids (Fix)}}	& \cellcolor[rgb]{0.93,0.43,0.41} 0.04	&\multicolumn{1}{c|}{\cellcolor[rgb]{0.94,0.54,0.52} 0.11}	&\cellcolor[rgb]{0.96,0.72,0.70} 0.22	&\multicolumn{1}{c|}{\cellcolor[rgb]{1.00,0.97,0.97} 0.38}	&\cellcolor[rgb]{0.92,0.38,0.36} 0.01	&\cellcolor[rgb]{0.70,0.81,0.61} 0.75 \\
\textit{LastBids (Var)}	& \cellcolor[rgb]{0.93,0.42,0.39} 0.03	&\cellcolor[rgb]{0.94,0.49,0.47} 0.08	&\cellcolor[rgb]{0.94,0.54,0.52} 0.11	&\cellcolor[rgb]{0.94,0.53,0.51} 0.10	&\cellcolor[rgb]{0.93,0.46,0.44} 0.06	&\cellcolor[rgb]{0.96,0.65,0.64} 0.18\\
\midrule
Avg & 0.13	&0.18 (+0.05)	&0.18	&0.35 (+0.17)	&0.26	&0.65 (+0.39)\\
\bottomrule
\end{tabular}}
\label{app:tab:open_winrate}
\end{minipage}
\end{table}

\section{Batter Nash Equilibrium Approximation with K-Level Reasoning}
\label{app:fasterEqui}
To assess the impact of K-Level Reasoning on Nash equilibrium, we conducted experiments testing the performance of LLMs at different K-Levels in the Prisoner's Dilemma.
The Prisoner's Dilemma is a classic concept in game theory that illustrates a paradoxical situation in which individuals acting in their own self-interest lead to a suboptimal outcome for everyone involved. It's often used to study decision-making in situations involving cooperation and competition.

From Table \ref{tab:prisoners-dilemma}, it is evident that when K=1, players exhibit weaker rationality: they tend to choose cooperation over betrayal. As the K-Level of one player increases (first row from left to right), players with higher K-Levels demonstrate stronger rationality: they tend to choose betrayal over cooperation. Moreover, when both players' K-Levels increase, they are more likely to reach a state of Nash equilibrium: both choosing to betray each other.
\renewcommand{\arraystretch}{1.5}

\begin{table}[ht]
\centering
\caption{Statistical analysis of the Payoff Matrix for LLM in the Prisoner's Dilemma across different K-Levels. The Nash equilibrium of the Prisoner's Dilemma, where both players choose to betray each other, is highlighted with a green background in the bottom-right corner of each cell.}
\resizebox{0.8\columnwidth}{!}{
    \begin{tabular}{lc:cc:cc:cc:c}
\hline
                                               & \multicolumn{2}{l}{Level-1}                         & \multicolumn{2}{l}{Level-2}                        & \multicolumn{2}{l}{Level-3}                         & \multicolumn{2}{l}{Level-4}                                 \\ \hline
\multicolumn{1}{l|}{}                          & 10 & \multicolumn{1}{c|}{0}                         & 7 & \multicolumn{1}{c|}{3}                         & 1 & \multicolumn{1}{c|}{9}                          & 0 & \multicolumn{1}{c|}{10}                                 \\
\cdashline{2-9}
\multicolumn{1}{l|}{\multirow{-2}{*}{Level-1}} & 0  & \multicolumn{1}{c|}{\cellcolor[HTML]{A6C089}0} & 0 & \multicolumn{1}{c|}{\cellcolor[HTML]{A6C089}0} & 0 & \multicolumn{1}{c|}{\cellcolor[HTML]{A6C089}0}  & 0 & \multicolumn{1}{c|}{\cellcolor[HTML]{A6C089}0}          \\ \cline{2-9} 
\multicolumn{1}{l|}{}                          & 7  & \multicolumn{1}{c|}{0}                         & 2 & \multicolumn{1}{c|}{0}                         & 0 & \multicolumn{1}{c|}{2}                          & 0 & \multicolumn{1}{c|}{1}                                  \\
\cdashline{2-9}
\multicolumn{1}{l|}{\multirow{-2}{*}{Level-2}} & 3  & \multicolumn{1}{c|}{\cellcolor[HTML]{A6C089}0} & 5 & \multicolumn{1}{c|}{\cellcolor[HTML]{A6C089}3} & 0 & \multicolumn{1}{c|}{\cellcolor[HTML]{A6C089}8}  & 0 & \multicolumn{1}{c|}{\cellcolor[HTML]{A6C089}9}          \\ \cline{2-9} 
\multicolumn{1}{l|}{}                          & 1  & \multicolumn{1}{c|}{0}                         & 0 & \multicolumn{1}{c|}{0}                         & 0 & \multicolumn{1}{c|}{0}                          & 0 & \multicolumn{1}{c|}{0}                                  \\
\cdashline{2-9}
\multicolumn{1}{l|}{\multirow{-2}{*}{Level-3}} & 9  & \multicolumn{1}{c|}{\cellcolor[HTML]{A6C089}0} & 2 & \multicolumn{1}{c|}{\cellcolor[HTML]{A6C089}8} & 0 & \multicolumn{1}{c|}{\cellcolor[HTML]{A6C089}10} & 1 & \multicolumn{1}{c|}{\cellcolor[HTML]{A6C089}9}          \\ \cline{2-9} 
\multicolumn{1}{l|}{}                          & 0  & \multicolumn{1}{c|}{0}                         & 0 & \multicolumn{1}{c|}{0}                         & 0 & \multicolumn{1}{c|}{1}                          & 0 & \multicolumn{1}{c|}{0}                                  \\
\cdashline{2-9}
\multicolumn{1}{l|}{\multirow{-2}{*}{Level-4}} & 10 & \multicolumn{1}{c|}{\cellcolor[HTML]{A6C089}0} & 1 & \multicolumn{1}{c|}{\cellcolor[HTML]{A6C089}9} & 0 & \multicolumn{1}{c|}{\cellcolor[HTML]{A6C089}9}  & 1 & \multicolumn{1}{c|}{\cellcolor[HTML]{A6C089}\textbf{9}} \\ \hline
\end{tabular}}
\label{tab:prisoners-dilemma}
\end{table}

\renewcommand{\arraystretch}{1}

\section{Computational Cost Analysis}
\label{app:computationalcost}
\begin{table}[ht]
\centering
\caption{Average input/output/total token consumption per game test. Unit: \textbf{Kilo} tokens.}
\tabcolsep=0.1cm
\resizebox{1\columnwidth}{!}{
\begin{tabular}{llrrrrrrr}
\toprule
\textbf{Game}                  & \textbf{Metric} & \textbf{Direct} & \textbf{CoT} & \textbf{Persona} & \textbf{Reflect} & \textbf{Refine} & \textbf{PCoT} & \textbf{K-R}  \\
\midrule
    \multirow{4}{*}{\textbf{GO8A}} & Input           & 14.80           & 23.30        & 18.70            & 39.50            & \textcolor{red}{\textbf{145.50}} & 30.80         & 123.40        \\
                               & Output          & 0.10            & 1.80         & 0.60             & 1.20             & \textcolor{red}{\textbf{6.90}}   & 1.30          & 1.60          \\
                               & Total           & 14.90           & 25.10        & 19.30            & 40.70            & \textcolor{red}{\textbf{152.40}} & 32.10         & 124.90 
                               \\ \cmidrule(lr){2-9}
                               & Avg Win Rate    & 0.16            & 0.42         & 0.41             & 0.24             & 0.37            & 0.40          & \textbf{0.65} \\
                               \midrule
\multirow{4}{*}{\textbf{SAG}}  & Input           & 21.30           & 29.00        & 27.10            & 51.40            & \textcolor{red}{\textbf{130.50}} & 37.00         & 90.10         \\
                               & Output          & 0.70            & 1.00         & 0.90             & 1.80             & \textcolor{red}{\textbf{3.30}}   & 1.30          & 1.40          \\
                               & Total           & 22.00           & 30.00        & 28.10            & 53.20            & \textcolor{red}{\textbf{133.80}} & 38.20         & 91.60   \\ \cmidrule(lr){2-9}
                               & Avg Surv Round  & 6.51            & 7.44         & 7.59             & 4.82             & 7.33            & 6.17          & \textbf{9.01}\\
                               \bottomrule
\end{tabular}}
\end{table}
To understand how different reasoning methods affect token consumption, thereby assisting users in utilizing models more effectively and optimizing resource utilization and cost control, we conducted an analysis of the computational expenditure associated with various reasoning approaches. Specifically, we calculated the average input/output/total token consumption per game test for both the "Guessing 0.8 of the Average" and "Survival Auction Game."

K-Level Reasoning, due to simulating the opponents' actions based on public historical behavior information with new sessions, inevitably causes more token consumption. The increase in token consumption linearly correlates with the number of opponents that need to be simulated and the rounds.

We also find that more token consumption does not necessarily lead to better results. The Self-Refine method, due to its action-feedback-refine pipeline, leads to a substantial increase in token consumption for both input and output, yet it did not outperform our proposed K-Level Reasoning in terms of performance. Compared to baselines, the proposed K-Level Reasoning method is an effective and efficient way to recursively simulate the decision-making process of the opponent and make decisions based on the simulation results.

\section{Statistical Significance of K-Level Reasoning efficacy}
\label{app:significance}

\begin{table*}[!ht]
\caption{The significance of metric result on G0.8A and SAG, comparing K-R against different \textbf{baseline models} in terms of Win Rate, Average Survival Round, and Adaptation Index.}
\resizebox{1\textwidth}{!}{
\begin{tabular}{@{}l|l|cccccc@{}}
\toprule
                    Game & Metric& Direct   & CoT      & Persona  & Reflect  & Refine   & PCoT          \\
\midrule
\multirow{2}{*}{G0.8A} &
Win Rate            & \textcolor{green!80!black}{\textbf{\checkmark}}(0.001) & \textcolor{green!80!black}{\textbf{\checkmark}}(0.005) & \textcolor{green!80!black}{\textbf{\checkmark}}(0.001) & \textcolor{green!80!black}{\textbf{\checkmark}}(0.001) & \textcolor{green!80!black}{\textbf{\checkmark}}(0.001) & \textcolor{green!80!black}{\textbf{\checkmark}}(0.001)        \\
&Adaption Index      & \textcolor{green!80!black}{\textbf{\checkmark}}(0.007) & \textcolor{green!80!black}{\textbf{\checkmark}}(0.010) & \textcolor{green!80!black}{\textbf{\checkmark}}(0.039) & \textcolor{red}{\textbf{$\times$}}(0.062) & \textcolor{green!80!black}{\textbf{\checkmark}}(0.003) & \textcolor{green!80!black}{\textbf{\checkmark}}(0.021)         \\
\midrule
\multirow{2}{*}{SAG} &
Average Survival Round            & \textcolor{green!80!black}{\textbf{\checkmark}}(0.003) & \textcolor{green!80!black}{\textbf{\checkmark}}(0.010) & \textcolor{red}{\textbf{$\times$}}(0.074) & \textcolor{green!80!black}{\textbf{\checkmark}}(0.001) & \textcolor{green!80!black}{\textbf{\checkmark}}(0.001) & \textcolor{green!80!black}{\textbf{\checkmark}}(0.001)        \\
&Adaption Index      & \textcolor{red}{\textbf{\textcolor{red}{\textbf{$\times$}}}}(0.337) & \textcolor{red}{\textbf{$\times$}}(0.786) & \textcolor{red}{\textbf{$\times$}}(0.659) & \textcolor{red}{\textbf{$\times$}}(0.066) & \textcolor{red}{\textbf{$\times$}}(0.672) & \textcolor{red}{\textbf{$\times$}}(0.894)        \\
\bottomrule
\end{tabular}}
\label{tab:sig-g08a}
\end{table*}

\begin{table*}[!ht]
\caption{The significance of comparing K-R with \textbf{PCoT} concerning the differential performance across various Baseline models on the metric of Prediction Accuracy.}
\resizebox{\textwidth}{!}{
\begin{tabular}{@{}l|ccccccc@{}}
\toprule
                    Game& Direct   & CoT      & Persona  & Reflect  & Refine   & PCoT     & K-R      \\
                    
\midrule
G0.8A & \textcolor{green!80!black}{\textbf{\checkmark}}(0.003) & \textcolor{red}{\textbf{$\times$}}(0.058) & \textcolor{green!80!black}{\textbf{\checkmark}}(0.031) & \textcolor{green!80!black}{\textbf{\checkmark}}(0.003) & \textcolor{green!80!black}{\textbf{\checkmark}}(0.022) & \textcolor{green!80!black}{\textbf{\checkmark}}(0.023) & \textcolor{red}{\textbf{$\times$}}(0.107) \\
SAG & \textcolor{red}{\textbf{$\times$}}(0.091) & \textcolor{green!80!black}{\textbf{\checkmark}}(0.023) & \textcolor{green!80!black}{\textbf{\checkmark}}(0.049) & \textcolor{red}{\textbf{$\times$}}(0.907) & \textcolor{red}{\textbf{$\times$}}(0.956) & \textcolor{red}{\textbf{$\times$}}(0.656) & \textcolor{green!80!black}{\textbf{\checkmark}}(0.028) \\
\bottomrule
\end{tabular}}
\label{tab:sig-sag}
\end{table*}

Due to the utilization of a large language model for evaluation, inherent stochasticity is present, and the sample size for testing is limited (only conducted 10 experiments). To objectively assess the reliability of the experiment result, we subjected the results from G0.8A and SAG to a t-test for significance with $p=0.05$. Specifically, different significance testing methods were employed for different evaluation metrics, as outlined below:

\begin{itemize}
    \item Win Rate \& Average Survival Round: A t-test was conducted on the Win Rate and Average Survival Round of K-R versus other baselines when facing different agents. The null hypothesis posited no significant difference in performance between K-R and other baselines concerning Win Rate \& Average Survival Round.

   \item Adaptation Index: The method akin to that of Win Rate \& Average Survival Round was applied. 

    \item  Prediction Accuracy: Since only PCoT and K-R explicitly predicted opponents, we compared the significance of the prediction differences between PCoT and K-R. Unlike Win Rate and Adaptation Index, we assessed the differences in predictions (averaged over multiple experiments) between PCoT and K-R when facing the same opponents (e.g., CoT) across different game rounds.
\end{itemize}

From Table \ref{tab:sig-g08a} and Table  \ref{tab:sig-sag}, it is evident that K-R exhibits a significant advantage over other baselines in Win Rate and Average Survival Round metrics. This indicates K-R's superior rationality in both G0.8A and SAG. Additionally, we observed weaker significance of K-R's Adaptation Index compared to baselines in SAG. We attribute this observation to: 1) the more factors considered by each agent in SAG (e.g., health status), resulting in a more dynamic and complex environment; and 2) the wider bidding range in later rounds of SAG games, potentially leading to greater bias in agent bidding and rendering the Adaptation Index a challenging metric.

\begin{figure*}[!ht]
  \centering
\includegraphics[width=1\textwidth]{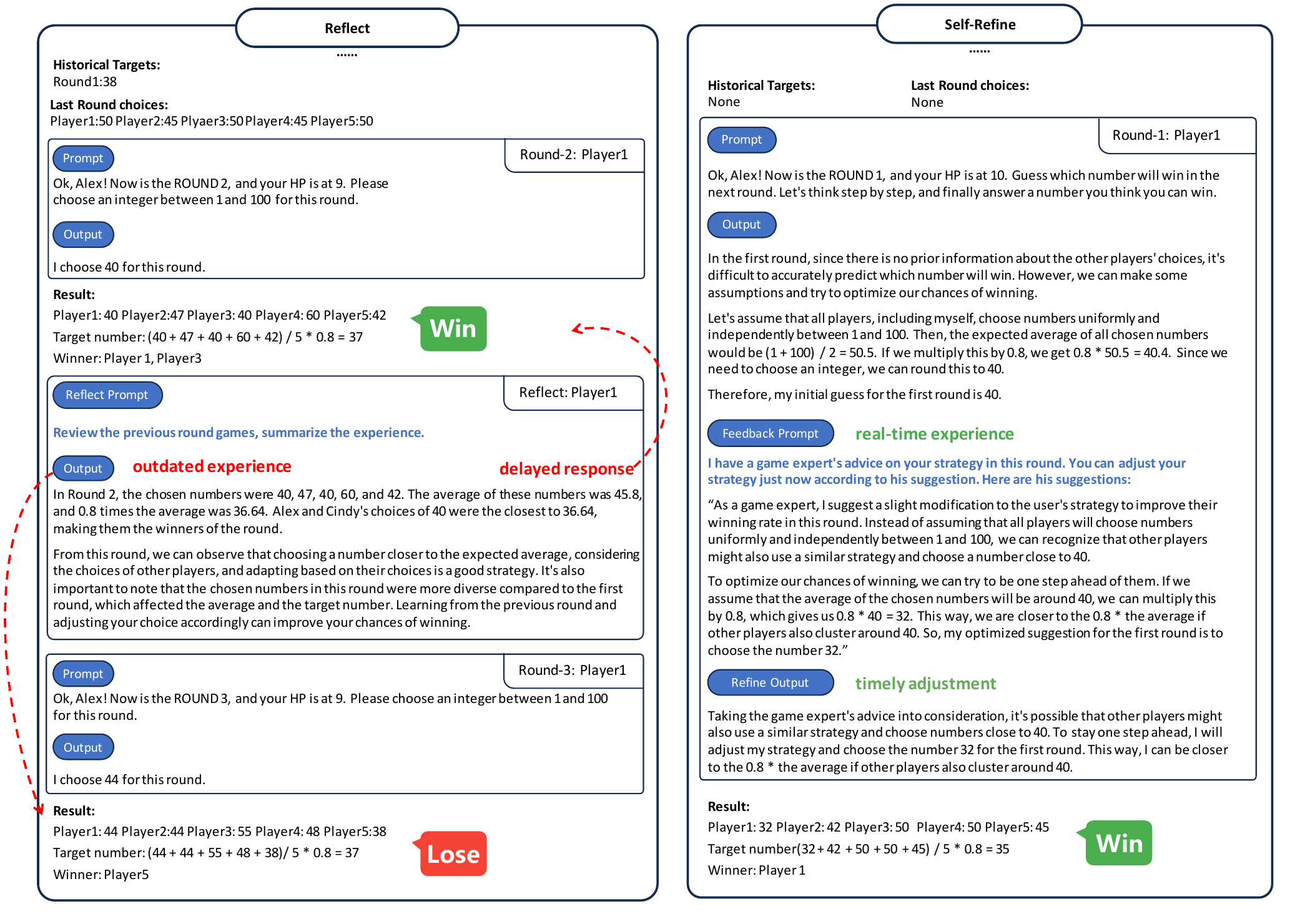}
  \caption{ Illustration of Reflect and Refine methods in the Guessing 0.8 of the Average game.}
\end{figure*}
\section{Timing in Dynamic Reasoning }
\label{app:time}

In our experiment, we implemented two LLM self-refinement reasoning methods: Reflect  \cite{madaan2023self} and Refine  \cite{shinn2023reflexion}, and noticed that Refine performed significantly better than Reflect in the experimental results. To further explore the differences in performance between these two methods, we analyzed their respective working principles and applicable scenarios.

The Reflect method involves making decisions first, then summarizing the experience based on feedback from the environment. This method may be effective in scenarios where the environment does not change much or where the decision-making cycle is long, as it allows for quick decision-making. However, its drawback is that, in dynamic environments, the experience from the previous round may not be suitable for the next round. In fact, in the Survival Auction Game (SAG), a rapidly changing environment, the survival rate of the Reflect method is even lower compared to making direct decisions. This is likely because this method does not sufficiently take into account the dynamic nature of the environment.

In contrast, the Refine method involves multiple analyses before making a decision, including an initial analysis and improvements to that initial analysis. Importantly, both of these analyses are conducted in the context of the current decision-making environment. This makes the Refine method more adaptable to dynamic environments, as it can consider real-time changes in the current environment, thus making more accurate decisions.

In summary, the reason why the Refine method performed better in our experiment is mainly that it adapts better to rapidly changing dynamic environments.

\section{Better Adaptability: Self-Refine vs. K-Level Reasoning}
\begin{figure}[ht]
\centering
  \includegraphics[width=1\columnwidth]{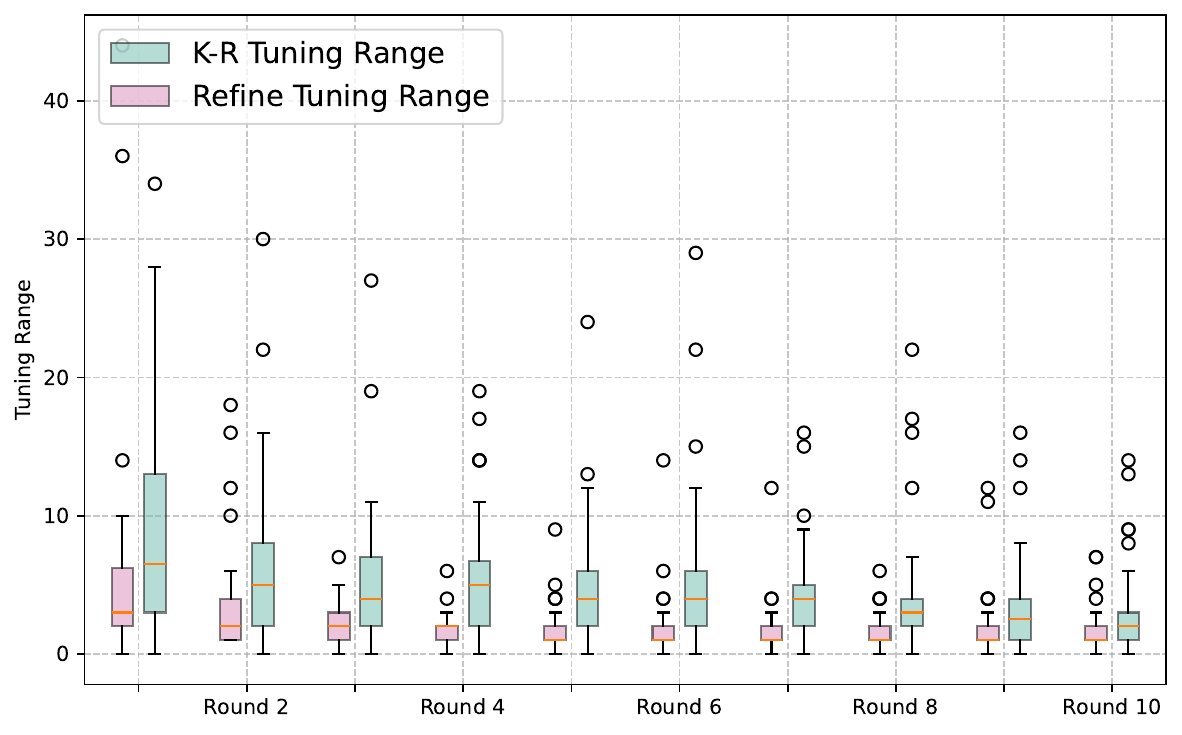}
\caption{The tuning range before and after adjustment during the G0.8A between PCoT and K-Level Reasoning. A larger tuning range indicates that the method exerts a more substantial influence.}
\label{fig:tuning_range}
\end{figure}
As described in Section \ref{sec:method_imp}, K-R optimizes initial actions through predictions of opponent behavior, thereby implementing higher-order beliefs. Although Self-Refine also optimizes initial actions via self-feedback, previous experiments have demonstrated that K-R significantly outperforms Self-Refine. To further investigate the source of K-R's performance enhancement, we present in Figure \ref{fig:tuning_range} a statistical analysis of action differences before and after optimization for both K-R and Self-Refine.

Analysis of the results reveals that Self-Refine initially exerts a substantial influence on actions. However, as time progresses, its optimization scope gradually diminishes, leading to a decrease in efficacy. In contrast, K-R exhibits a broader range of optimization, indicating that the higher-order belief mechanism has a more pronounced impact on K-R's performance.

This disparity primarily stems from the distinct feedback mechanisms employed by K-R and Self-Refine. K-R generates feedback based on predictions of opponent behavior, enabling it to capture trends in opponent actions and form an understanding of the environment, thus facilitating rapid adaptation. Conversely, Self-Refine's feedback is derived from its own strategy, which often adheres to fixed patterns. Consequently, Self-Refine underperforms K-R in terms of environmental adaptability.

\section{Reasoning Method Implementation Details}
\label{app:prompt}
\begin{figure*}[ht]
  \centering
\includegraphics[width=1\textwidth]{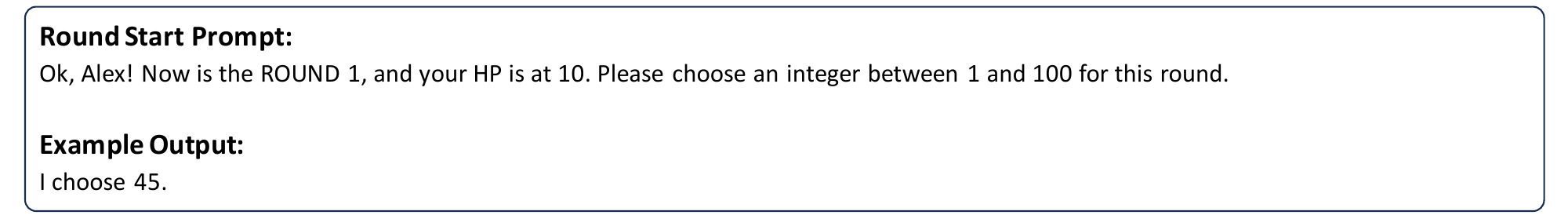}
  \caption{G0.8A: Direct Prompt}
\end{figure*}
\label{app:reflect}

\begin{figure*}[ht]
  \centering
\includegraphics[width=1\textwidth]{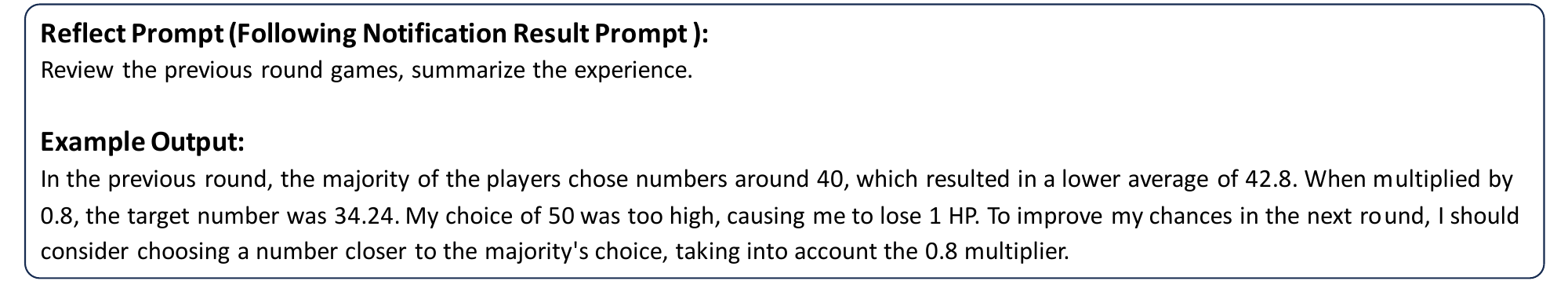}
  \caption{G0.8A-Reflect Prompt: Because the Reflect method requires correct labels, which is unrealistic in dynamic problems, we modified this method to have the LLM attempt to summarize the experience of victory or failure at the end of each round, and save the generated feedback content to assist the decision-making process in the next round. The content in Appendix \ref{app:time} shows the implementation differences between Reflexion and Self-Refine.}
\end{figure*}
\begin{figure*}[ht]
  \centering
\includegraphics[width=1\textwidth]{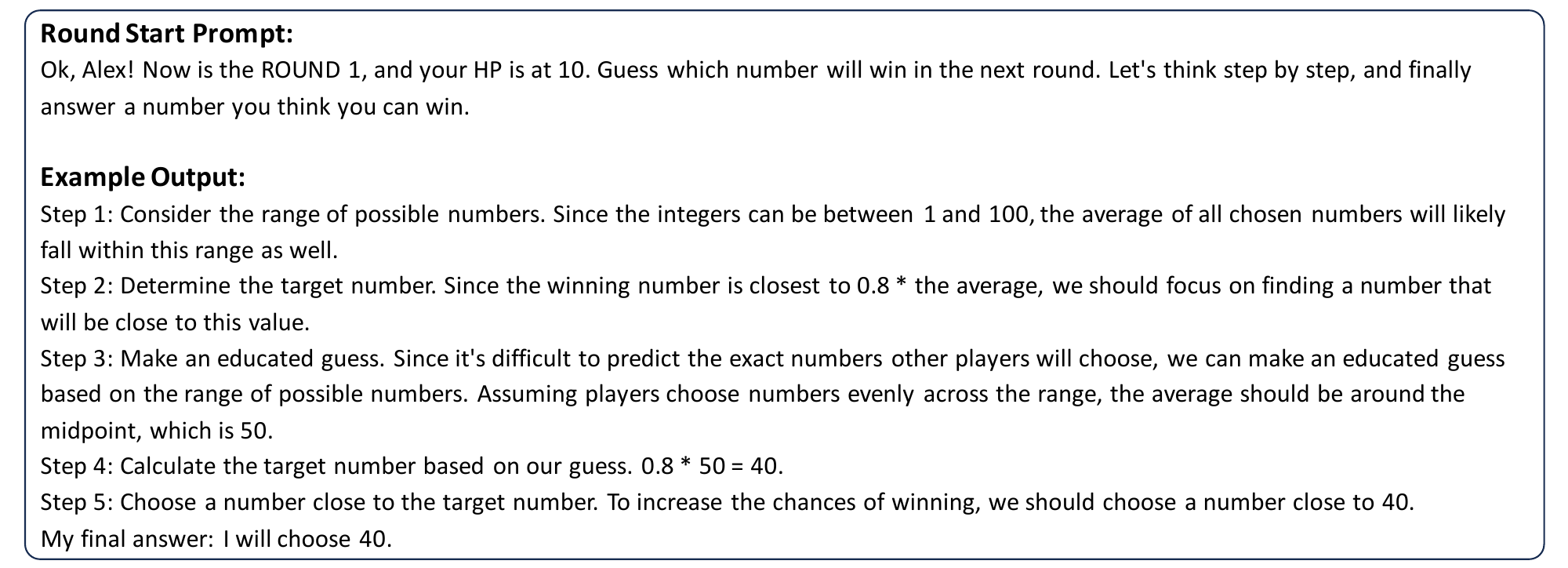}
\caption{G0.8A: CoT Prompt}
\end{figure*}
\label{app:pcot}
\begin{figure*}[ht]
  \centering
\includegraphics[width=1\textwidth]{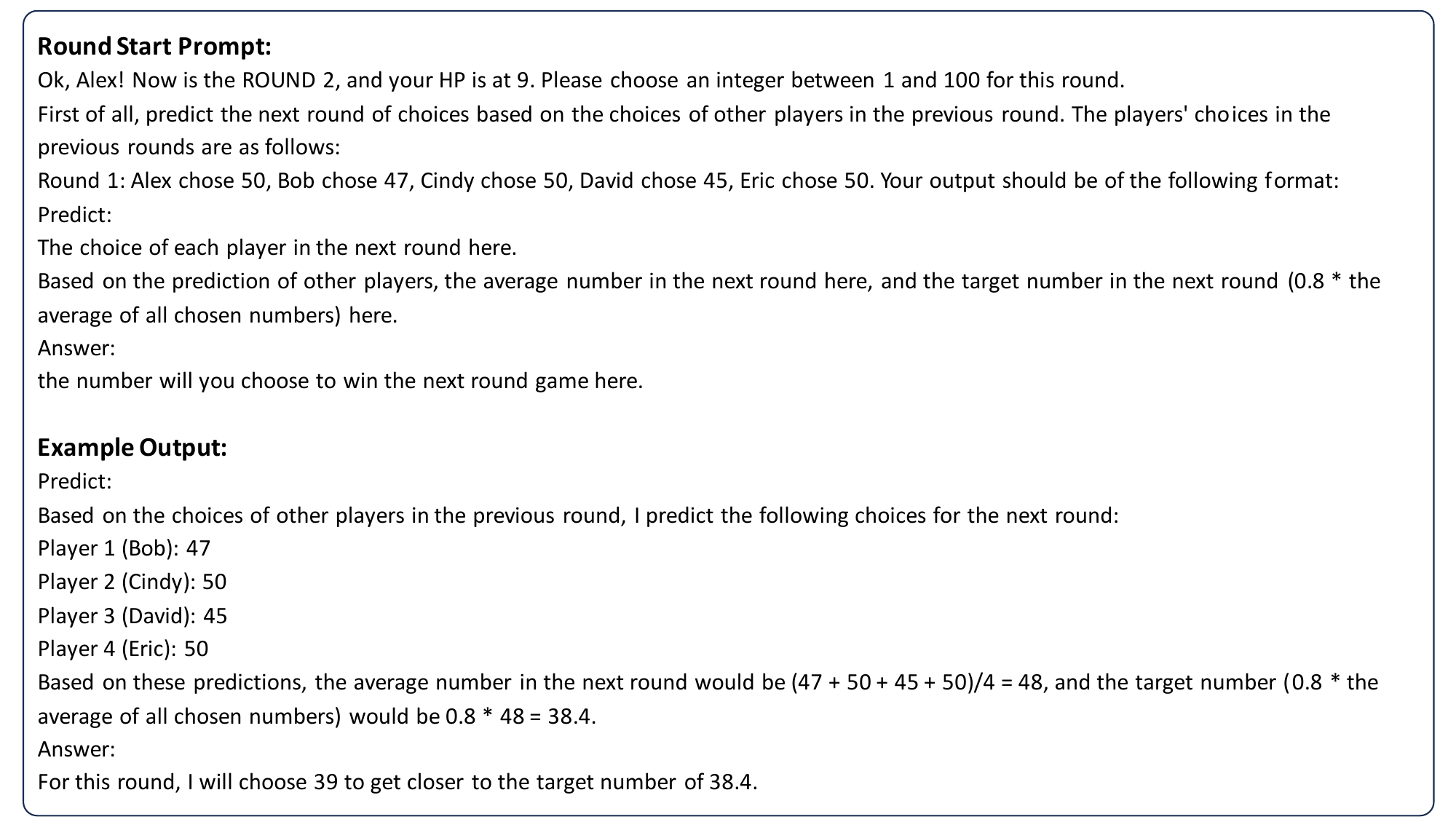}
\caption{G0.8A PCot Prompt: PCot diverges from the standard Chain of Thought (CoT) by \textbf{explicitly} requiring the LLM to \textbf{predict opponents' actions} before making a decision. This method responds to the immediate problem by anticipating future scenarios, which is crucial in dynamic reasoning.}
\end{figure*}
\begin{figure*}[ht]
  \centering
\includegraphics[width=1\textwidth]{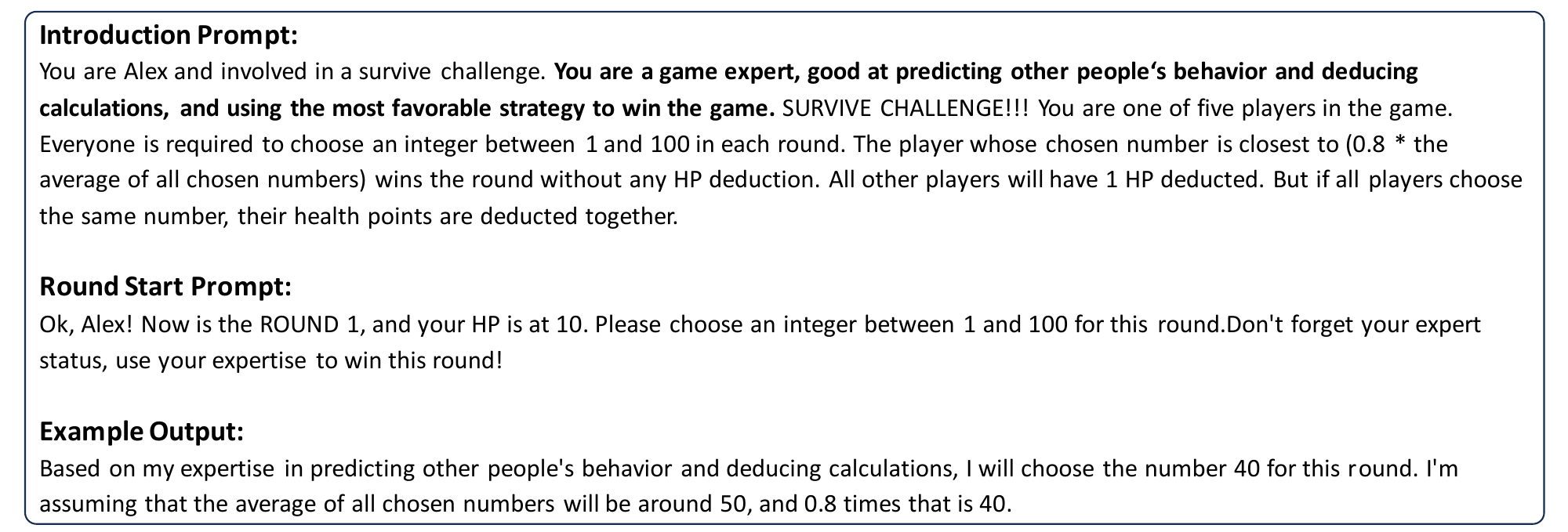}
\caption{G0.8A:Persona Prompt}
\end{figure*}
\begin{figure*}[ht]
  \centering
\includegraphics[width=1\textwidth]{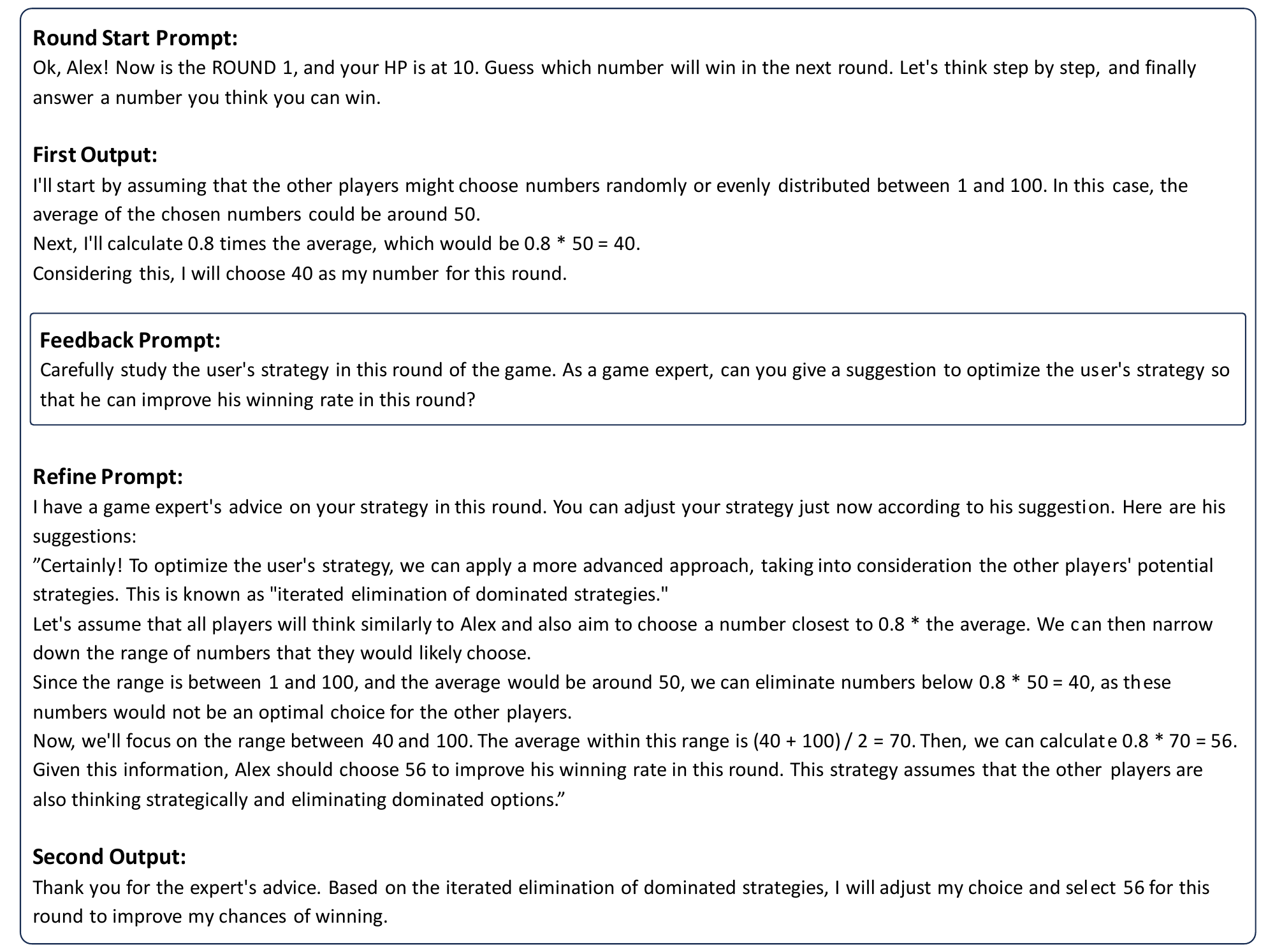}
\caption{G0.8A: Refine Prompt}
\end{figure*}
\begin{figure*}[ht]
  \centering
\includegraphics[width=1\textwidth]{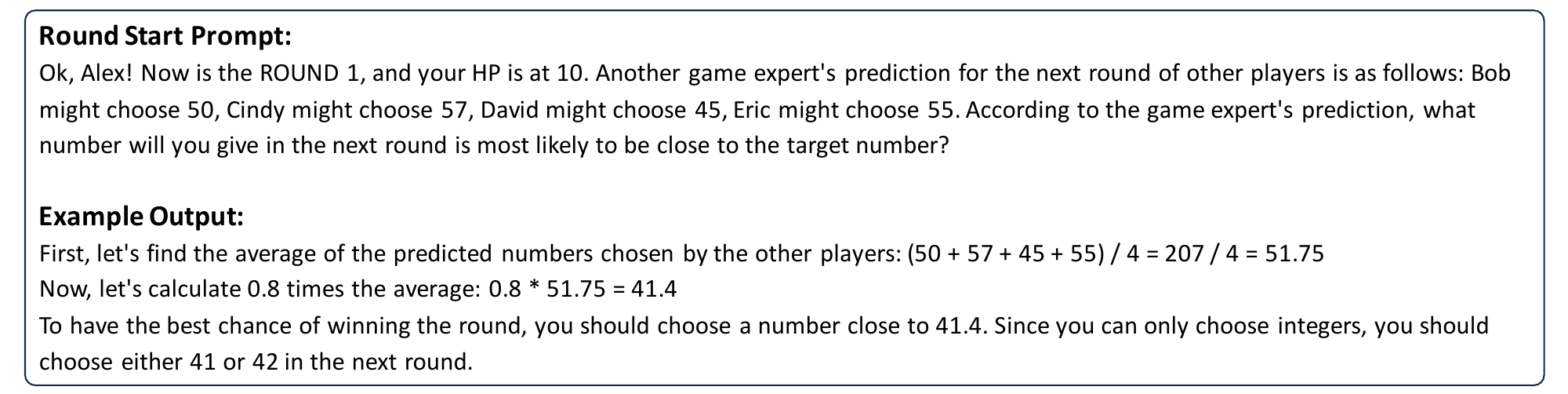}
\caption{G0.8A: K-Level Reasoning}
\end{figure*}

\begin{figure*}[ht]
  \centering
\includegraphics[width=1\textwidth]{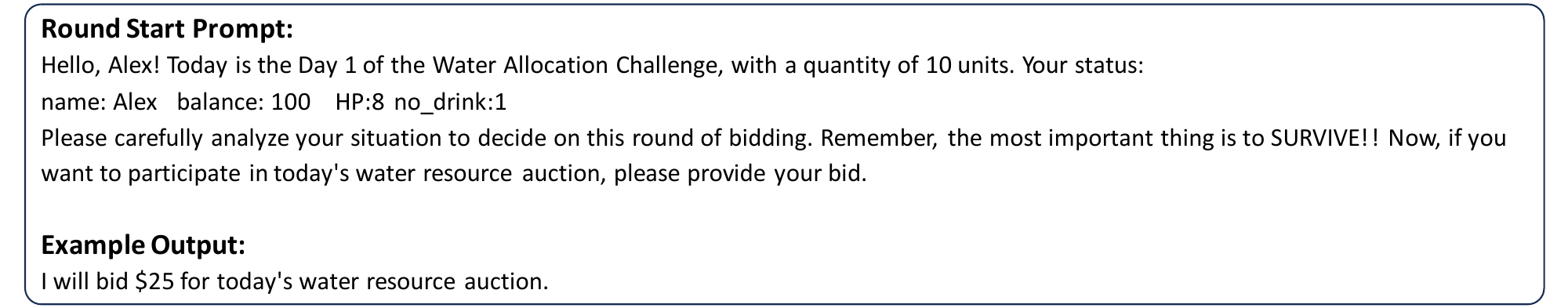}
\caption{SAG: Direct Prompt}
\end{figure*}

\begin{figure*}[ht]
  \centering
\includegraphics[width=1\textwidth]{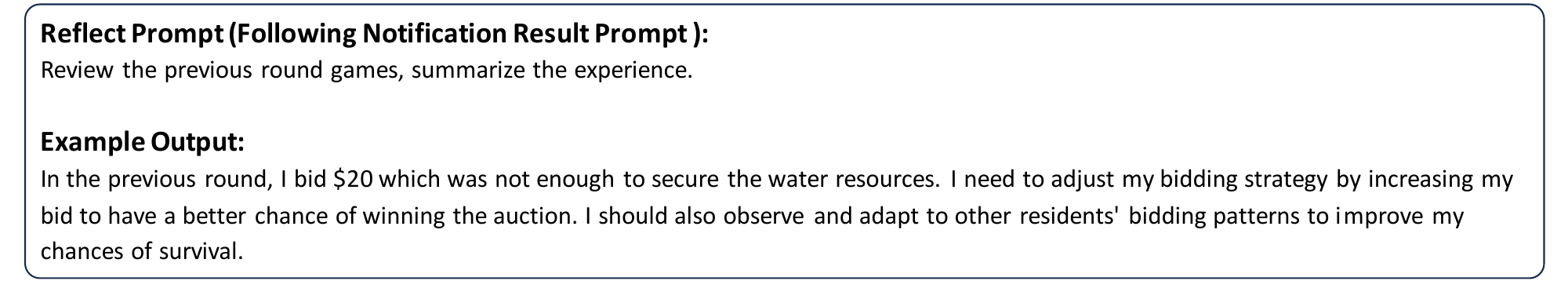}
\caption{SAG: Reflect Prompt}
\end{figure*}

\begin{figure*}[ht]
  \centering
\includegraphics[width=1\textwidth]{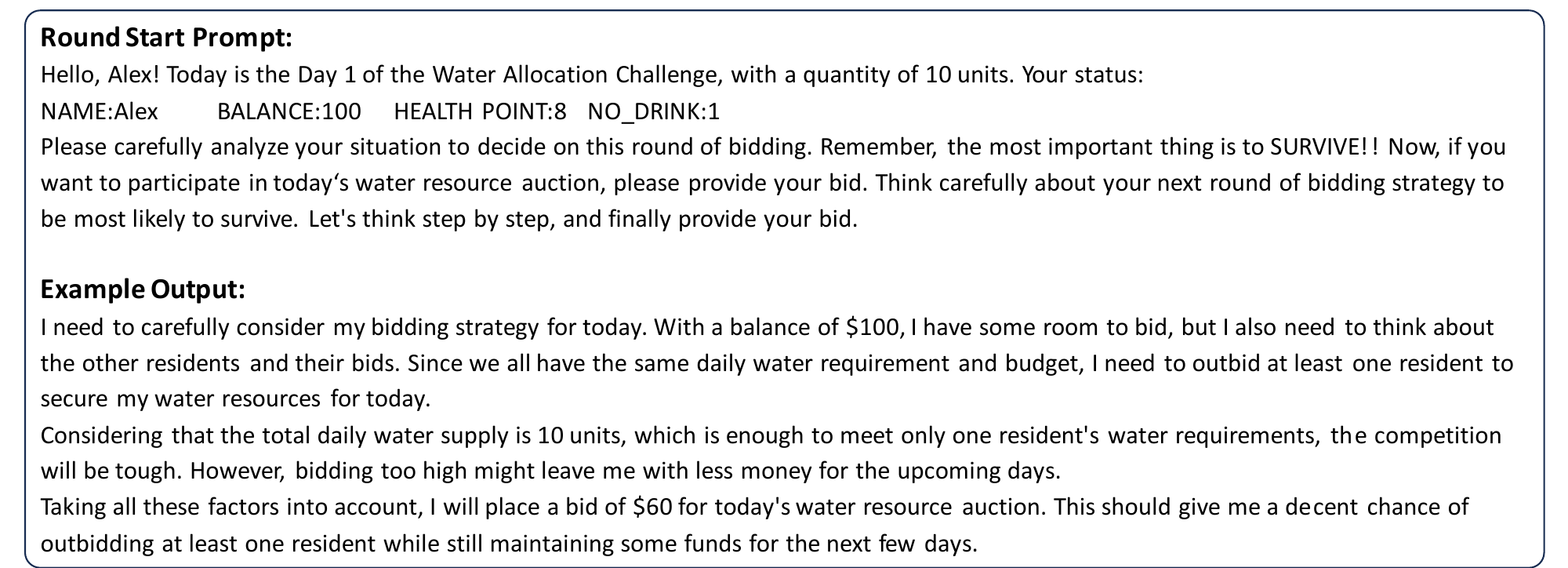}
\caption{SAG: CoT Prompt}
\end{figure*}
\begin{figure*}[ht]
  \centering
\includegraphics[width=1\textwidth]{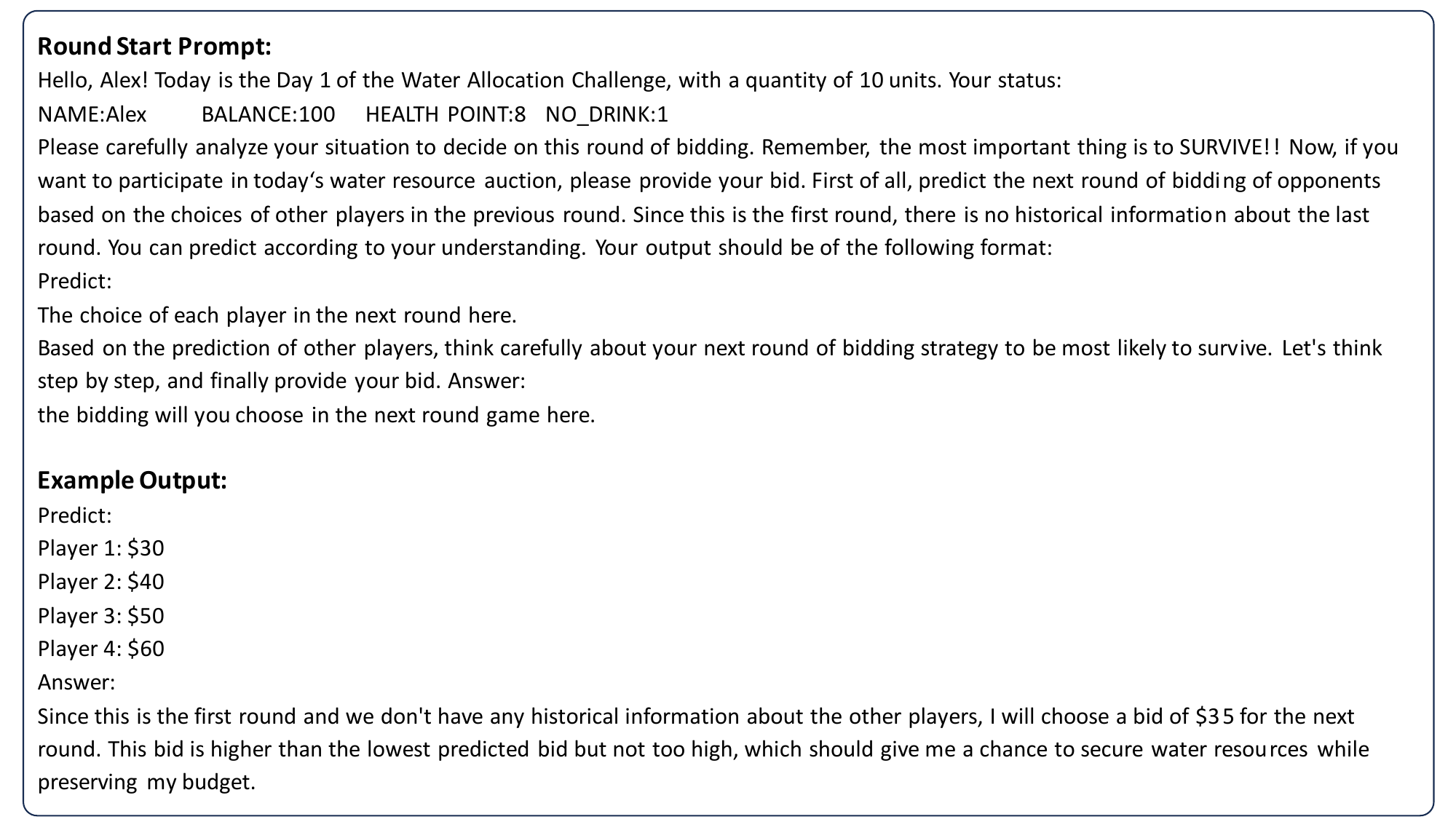}
\caption{SAG: PCoT Prompt}
\end{figure*}
\begin{figure*}[ht]
  \centering
\includegraphics[width=1\textwidth]{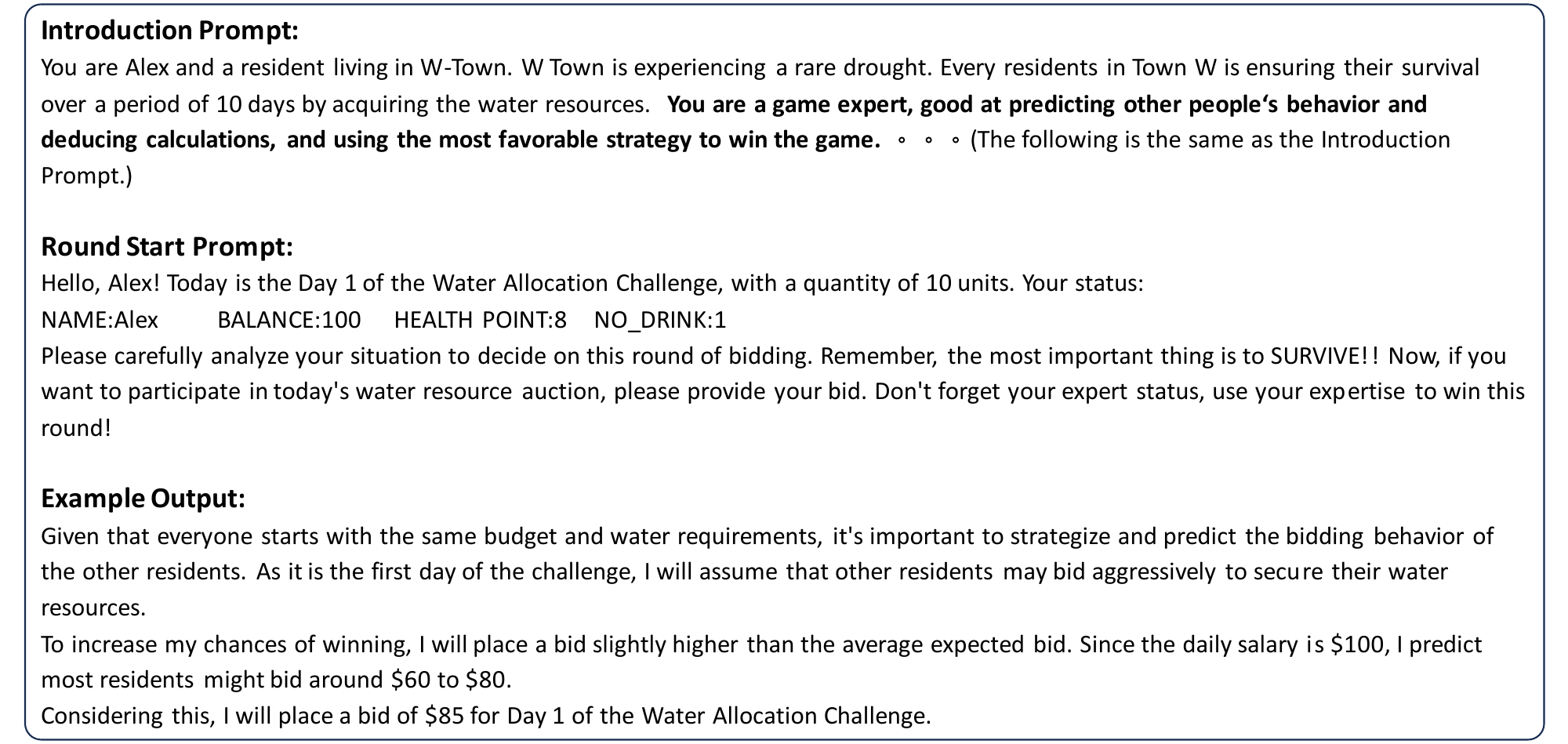}
\caption{SAG: Persona Prompt}
\end{figure*}
\begin{figure*}[ht]
  \centering
\includegraphics[width=1\textwidth]{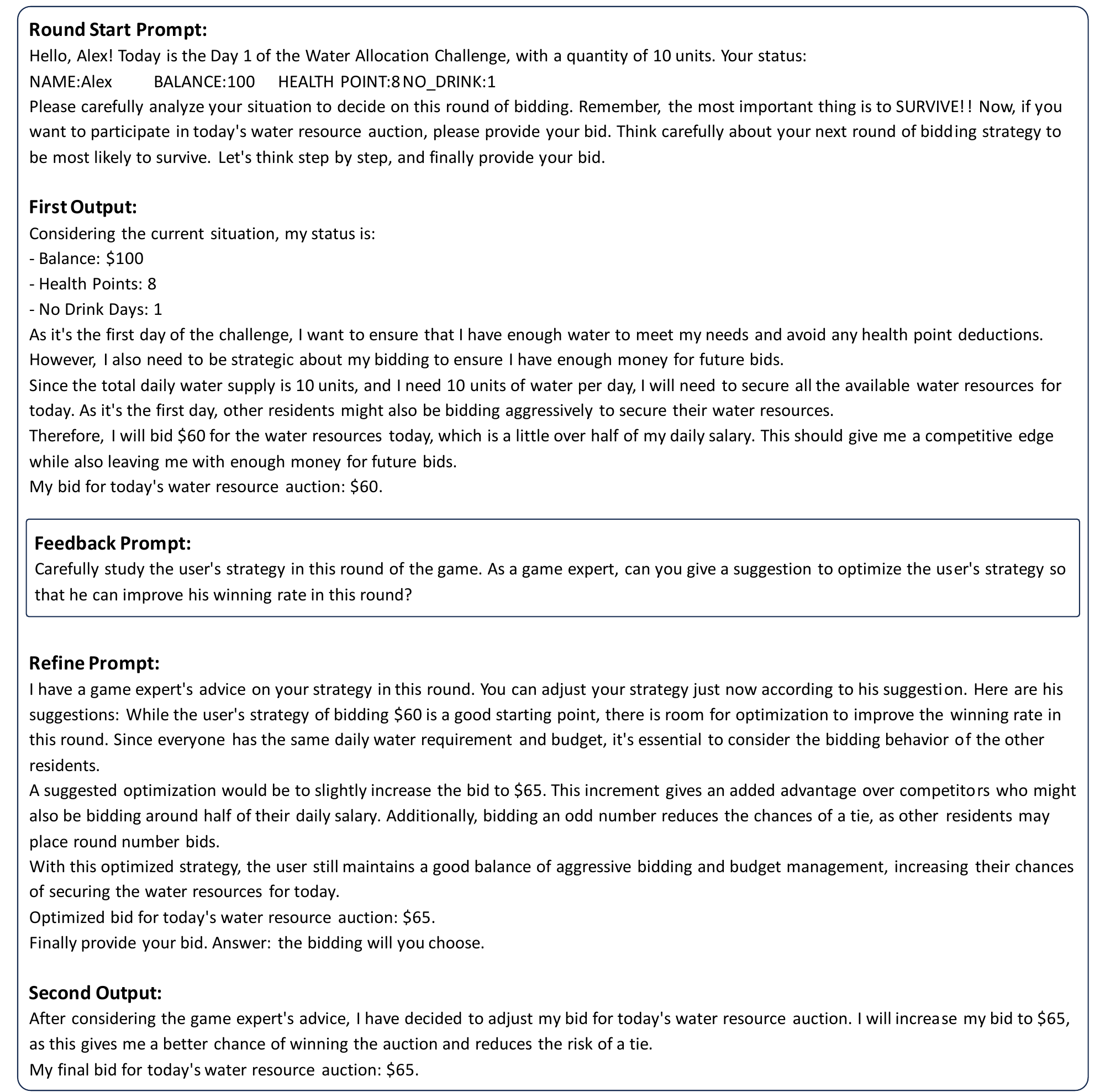}
\caption{SAG: Refine Prompt}
\end{figure*}
\begin{figure*}[ht]
  \centering
\includegraphics[width=1\textwidth]{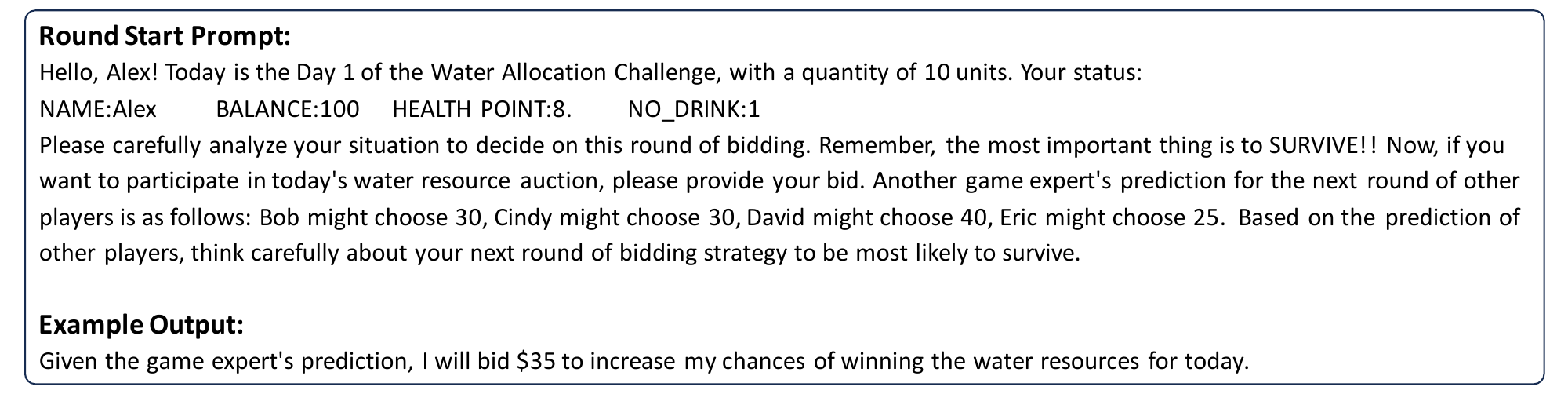}
\caption{SAG: K-Level Reasoning Prompt}
\end{figure*}

\end{document}